%% file: main.tex
\documentclass[11pt]{article}
\usepackage[preprint]{acl}
\usepackage{times}
\usepackage[T1]{fontenc}
\usepackage[utf8]{inputenc}
\usepackage{microtype}
\usepackage{booktabs}
\usepackage{amsmath}
\usepackage{graphicx}
\usepackage{enumitem}
\usepackage{amsmath}
\makeatletter
\@ifpackageloaded{lineno}{%
  \IfFormatAtLeastTF{2025-06-01}{%
    \RemoveFromHook{build/column/before}[lineno]%
    \AddToHook{build/column/after}[aclfix]{%
      \setbox\@outputbox\vbox{%
        \boxmaxdepth\@maxdepth
        \protected@write\@auxout{}{%
          \string\def\string\@LN@column{\if@firstcolumn1\else2\fi}}%
        \box\@outputbox}}%
  }{}%
}{}
\makeatother

\newcommand{\bench}{PreRun}

\title{How Long Will My Language Model Run?\\A Benchmark for Pre-Execution Runtime Prediction}
\author{
Ashish Molakalapalli\textsuperscript{1},
Kangda Wei\textsuperscript{1},
Haonan Duan\textsuperscript{2},\\
\textbf{Alex Carter\textsuperscript{2}},
\textbf{Ruihong Huang\textsuperscript{1}}\\
\textsuperscript{1}Texas A\&M University, College Station, TX\\
\textsuperscript{2}XPerf Inc, Round Rock, TX\\
\texttt{\{ashish\_molakalapalli,kangda,huangrh\}@tamu.edu}\\
\texttt{\{haonan.duan,alex.carter\}@xperf.ai}
}
\begin{document}
    \maketitle
\begin{abstract}
Researchers submitting deep learning jobs to shared GPU clusters need runtime estimates to request time limits, choose hardware, and budget compute. Large language models (LLMs) offer a tempting shortcut: describe the job and ask. We introduce \bench{}, a benchmark for predicting deep learning training and inference runtime from information available before execution. \bench{} contains 801 training and 1,884 inference records of language and vision models collected on a GPU cluster, paired with workload configurations and explicitly defined timing targets. It also includes 64 text-generation configurations measured under standardized conditions, each in three separate program launches. We compare three LLMs (Qwen3.5-9B, GPT-OSS-120B, and Claude Opus 5.5) with regression baselines and evaluate analytical estimates on a supported forward-pass subset. On 16 held-out generation configurations, the open-weight LLMs underestimate latency by 90--99\% on average without examples. Claude's zero-shot mean absolute error is 4.5 times that of a Random Forest fit on 32 configurations. Five labeled examples substantially reduce Claude's generation error, but their benefit varies across timing targets. We will open-source our data and evaluation pipeline upon acceptance.
\end{abstract}
\begin{figure*}[t]
\centering
\includegraphics[width=\textwidth]{{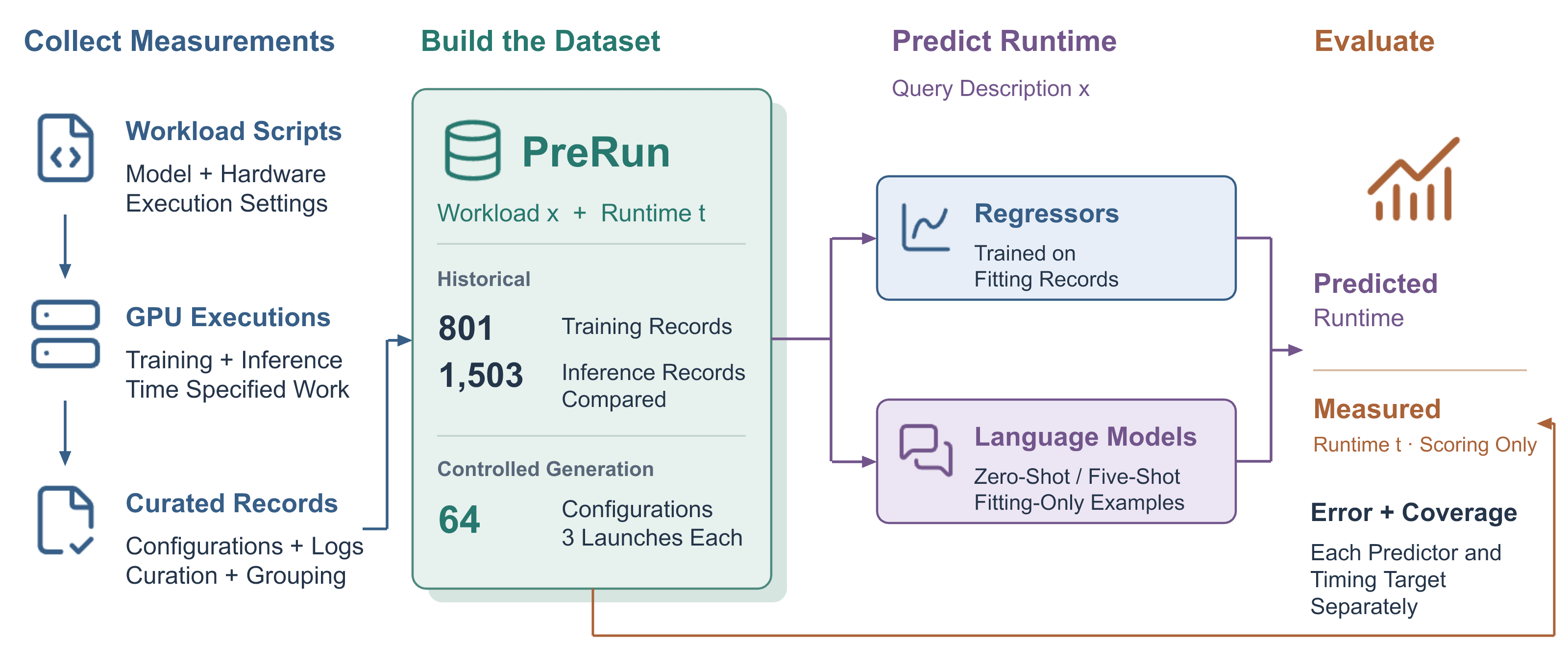}}
\vspace{-0.8cm}
\caption{Overview of \bench{}. GPU measurements produce workload--runtime records. Predictors receive pre-execution descriptions; fitting records supply regressor training and five-shot examples. Measured query runtimes enter scoring only. Error and coverage are evaluated separately for each timing target and collection.}
\label{fig:benchmark-overview}
\end{figure*}
\input{introduction}

\input{related}

\input{task}
\input{dataset}
\input{evaluation}
\input{results}
\section{Conclusion}

We introduced \bench{}, a benchmark for predicting deep learning
training and inference runtime before execution, with explicitly
defined timing targets. Our experiments show that prediction
accuracy depends on the workload, predictor, and examples
provided. Five examples substantially improve Claude's generation
estimates, but their benefit is not consistent across timing
targets. Simple regressors remain strong baselines, while
open-weight LLMs have large generation errors in the tested
settings. We hope \bench{} will support predictors that combine
the broad knowledge of LLMs with measured data, and we plan to
extend it to more hardware, workload models, and natural prompts.

\section*{Limitations}

The controlled collection covers two small workload models on
one GPU type in FP32, with shared synthetic token inputs.
Natural prompts may change output lengths and predictor
rankings. The historical collection comes from a single cluster
and was filtered partly by run outcomes
(Section~\ref{sec:collection}); it does not represent all
submitted jobs. Historical results use development records.
The controlled holdout contains only 16 configurations of known
models and hardware, and prior collection-quality checks limit
its independence. It is not external confirmation; small
differences require interpretation alongside uncertainty
intervals. LLM results reflect specific versions, fixed
examples, and unequal inference budgets. Alternative examples, reasoning budgets, retrieval, and tool use
remain untested. \newpage
\bibliography{references}
\appendix
\input{appendix}

\end{document}

%% file: introduction.tex
\section{Introduction}

Deep learning models are increasingly trained and served on shared
GPU clusters
\citep{jeon2019analysis,hu2021characterization,weng2022mlaas,hu2024characterization}.
Before submitting a job, a user has to make decisions that depend
on how long it will run: what time limit to request, which GPUs
to use, how large a batch to use, and whether a more expensive
decoding strategy such as beam search is affordable.
Getting the estimate wrong is costly. A job whose time limit is
too short can be killed before it finishes, while requesting too
much time can delay scheduling. User-provided runtime estimates
are known to be inaccurate, and better predictions can improve
scheduling \citep{mualem2001utilization,tsafrir2007backfilling}.
Many deep learning schedulers likewise rely on estimates of job
duration or throughput
\citep{peng2018optimus,narayanan2020gavel,hu2023lucid}.
As the compute cost of NLP research grows, such estimates also
matter for budgeting and energy use
\citep{strubell2019energy,schwartz2020green}.

There are several ways to estimate runtime in advance.
Analytical models relate the floating-point operations and memory
traffic of a workload to hardware throughput
\citep{williams2008roofline,qi2017paleo}.
Profiling-based predictors measure part of a workload and
extrapolate, sometimes using learned models
\citep{zhu2020daydream,yu2021habitat}.
These approaches can require systems expertise or access to
measurement hardware. A simpler option is to describe the job
to a large language model (LLM) and ask how long it will take.
LLMs can perform regression from in-context examples
\citep{vacareanu2024words,song2024omnipred}, and recent systems
use them to predict job runtimes on HPC platforms
\citep{liu2025ora} and inform GPU cluster scheduling
\citep{wang2025schedmate}. This raises a practical question:
how accurate are their estimates across deep learning workloads
and timing targets?

Answering this question requires detailed workload descriptions
paired with clearly defined measurements. The cluster traces
considered here record job activity and resource use but lack
the full configuration, such as batch size, sequence lengths,
and decoding settings, needed for these comparisons
\citep{jeon2019analysis,hu2021characterization,weng2022mlaas}.
MLPerf and DAWNBench evaluate reference workloads
\citep{coleman2017dawnbench,mattson2020mlperf,reddi2019mlperf},
while datasets for learned cost models measure tensor programs
or computational graphs
\citep{zheng2021tenset,phothilimthana2023tpugraphs}.
A runtime benchmark must also specify exactly what is timed.
A single forward pass and a full \texttt{generate()} call on
the same model are different targets: generation includes
repeated model computations and decoding logic. Accuracy on
one therefore need not imply accuracy on the other.
A recent survey identifies richer workload data and comparable
evaluation as open challenges for runtime prediction
\citep{wei2026survey}.

We introduce \bench{}, a benchmark for predicting the runtime
of deep learning workloads before execution. Our earlier
measurement campaigns provide 801 training and 1,884 inference
records of language and vision models; we refer to these as
the historical collection. Each record pairs a workload
configuration with a measured duration whose scope is explicitly
defined. To compare generation settings using a consistent
timing procedure and repeated measurements, we also collect
64 text-generation configurations of Qwen2.5-0.5B and T5-small,
each in three separate program launches. This controlled
collection measures complete generation calls under standardized
conditions. We evaluate three LLMs (Qwen3.5-9B, GPT-OSS-120B,
and Claude Opus 5.5) with and without five labeled examples,
alongside regression and nearest-neighbor baselines.
A separate forward-pass study evaluates analytical estimates
and learned corrections.

Our experiments lead to three main findings:
\begin{itemize}[noitemsep, topsep=0pt]
\item \textbf{Zero-shot accuracy depends on the predictor
and target.}
On 16 held-out generation configurations, Qwen3.5-9B and
GPT-OSS-120B underestimate latency by 99\% and 90\% on average.
Claude Opus 5.5 is far more accurate, but its mean absolute
error (MAE) is still 4.5 times that of a Random Forest fit
on 32 configurations. On custom training loops, however,
zero-shot Claude has the lowest observed percentage error.

\item \textbf{Examples help, but unevenly.}
Even with five examples, the open-weight LLMs have mean
absolute percentage errors above 50\% on held-out generation
configurations. Claude benefits more: five examples reduce
its MAE on training time reported by Hugging Face Trainer
by 65\%. Across the six historical targets, its MAE improves
on four, but increases on custom training loops and slightly
on forward-pass prediction.

\item \textbf{Simple regressors provide strong baselines.}
The best method depends on the target and metric.
On causal LM and encoder--decoder forward sums, LightGBM's
MAE is 39--72\% lower than that of Claude with five examples.
On the controlled holdout, Random Forest has the lowest
observed MAE, although its paired comparison with five-shot
Claude does not establish a clear ordering.
\end{itemize}

%% file: related.tex
\section{Related Work}

\paragraph{Runtime prediction for deep learning jobs.}
Trace studies of production GPU clusters, including Microsoft's
Philly \citep{jeon2019analysis,phillytrace}, SenseTime's Helios
\citep{hu2021characterization}, Alibaba PAI
\citep{weng2022mlaas}, and clusters used for LLM development
\citep{hu2024characterization}, report job durations, GPU
allocations, and utilization. Schedulers use runtime information
in different ways: Optimus builds online models of training
progress \citep{peng2018optimus}, Gavel relies on per-accelerator
throughput estimates \citep{narayanan2020gavel}, Helios and Lucid
predict job durations from historical jobs
\citep{hu2021characterization,hu2023lucid}, and Tiresias is
designed for the case where durations are unknown
\citep{gu2019tiresias}. Language models have recently entered
this loop: ORA predicts HPC job runtimes with a
retrieval-augmented language model \citep{liu2025ora}, and
SchedMate uses LLMs to extract information from source code,
logs, and historical jobs for GPU scheduling
\citep{wang2025schedmate}. \bench{} complements these traces
and systems with per-run configurations and explicitly defined
timing targets.

\paragraph{Performance models and benchmarks.}
Analytical models estimate execution time from computation,
memory traffic, and hardware capabilities, as in Paleo
\citep{qi2017paleo} and the roofline model
\citep{williams2008roofline}. For LLM inference, analytical
and simulation models capture the different costs of prefill
and decoding
\citep{yuan2024llm,agrawal2024vidur,zhang2024llmcompass,patwari2025life}.
Profiling-based predictors use measured execution information
to estimate performance under changed configurations or hardware
\citep{zhu2020daydream,yu2021habitat}. NeuSight instead learns
tile-level performance models to predict latency on unseen GPUs
without executing the query workload there
\citep{lee2025neusight}.
Because generation time depends on output length, several
systems also predict response length before generation
\citep{jin2023s3,zheng2023response}; in \bench{}, the requested
output limit is an input but the realized length is not.
Relevant datasets and benchmarks include measurements of
individual tensor programs and graphs
\citep{chen2018learning,zheng2021tenset,phothilimthana2023tpugraphs}
and reference workloads
\citep{coleman2017dawnbench,mattson2020mlperf,reddi2019mlperf}.
\citet{wei2026survey} survey runtime prediction methods and
datasets for LLM training and inference.

%% file: task.tex
\section{Task}
\label{sec:task}

Given a description $x$ of a workload and a timing target $p$,
a predictor outputs a positive duration $\hat y=f_p(x)$ before
the workload runs. The timing target specifies which part of
execution is measured (Section~\ref{sec:historical-time});
training targets are in seconds and inference targets in
milliseconds. We evaluate each target separately within each
collection and never pool errors across targets.

\paragraph{Inputs.}
The description contains what a user knows before launching
a job: the model, the requested work (batch size, input length
or image size, epochs or steps, and requested output limit),
the hardware (GPU type and count), and execution settings
such as precision and decoding strategy. For historical runs,
these fields are recovered from each run's launch configuration
and script; Appendix~\ref{app:schemas} lists them.
Run identifiers, file paths, and split assignments are never
used as features.

\paragraph{Excluded information.}
Outcomes of the query workload are excluded from predictor
inputs: its runtime, the number of tokens actually generated,
whether generation stopped early, measured memory use, and
GPU utilization. In particular, the requested output limit
is an input, but the realized output length is not.
Measured runtimes from fitting records remain available
for learning, lookup, and examples.

\paragraph{Terminology.}
We call the models being timed \emph{workload models} and
the LLMs that make predictions \emph{predictor LLMs}.
In the controlled collection, for example, Qwen2.5-0.5B
is a workload model, whereas Qwen3.5-9B is a predictor LLM.

%% file: dataset.tex
\section{The \bench{} Dataset}
\label{sec:data}

\begin{figure}[!t]
\centering
\includegraphics[width=\columnwidth]{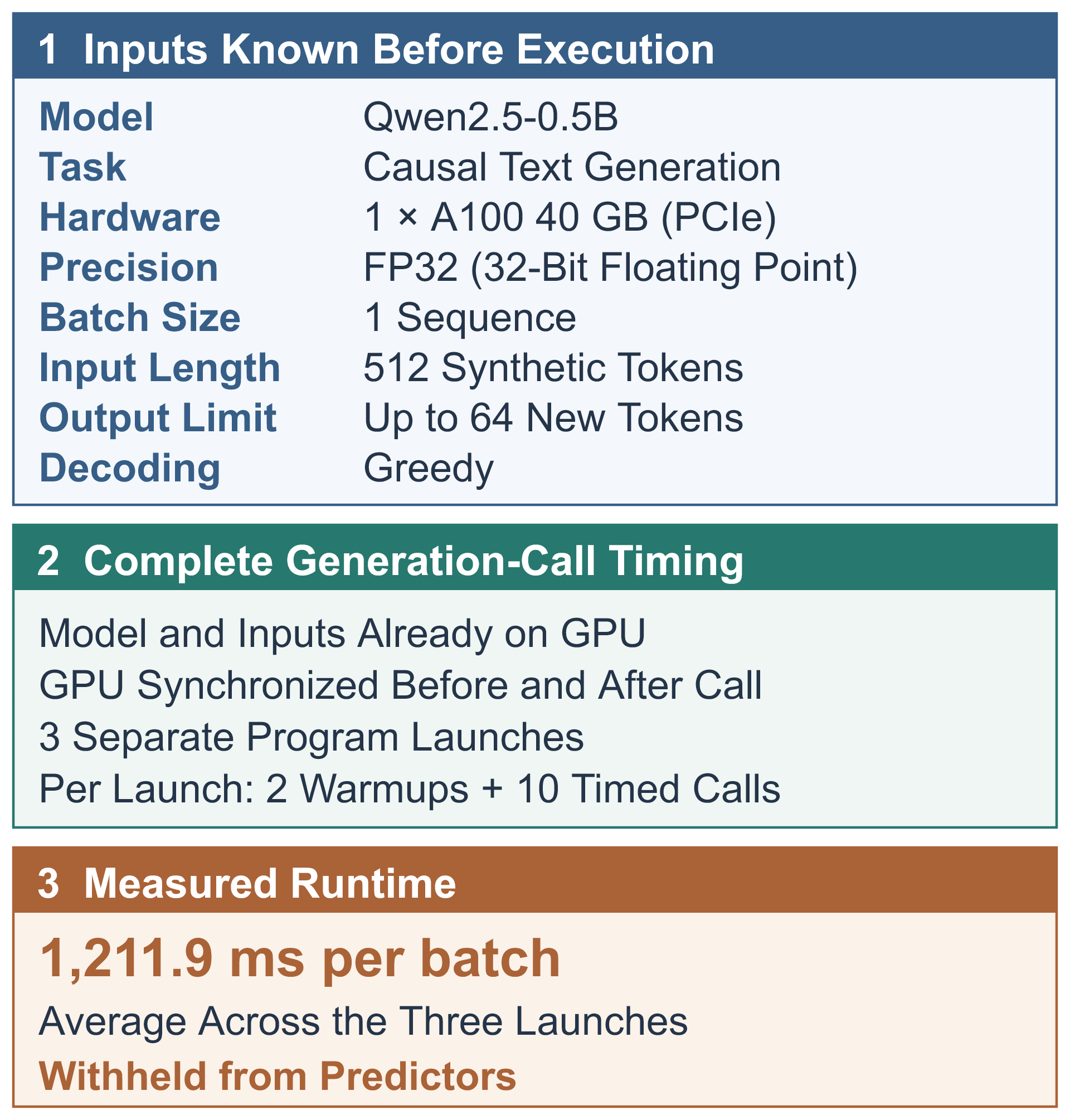}
\caption{An actual development example from controlled generation, using synthetic token inputs. Each launch excludes two warmups and averages ten timed calls. The reported runtime averages the three launch means; generation may stop before the output limit.}
\label{fig:data-example}
\end{figure}

\bench{} pairs workload descriptions with measured runtimes.
We assembled 801 training and 1,884 inference records from our
earlier GPU measurement campaigns, termed the \emph{historical
collection}. We also measured 64 text-generation configurations
repeatedly under a common procedure, termed the \emph{controlled
generation collection}. ``Controlled'' describes how measurements
were collected, not a type of evaluation split.
Table~\ref{tab:population-overview} summarizes the records used
in our comparisons; Section~\ref{sec:data-splits} explains their roles. Figure~\ref{fig:data-example} illustrates one controlled-generation
configuration, its timing procedure, and the measured runtime
that predictors aim to estimate.

\begin{table*}[t]
\centering
\small
\begin{tabular}{lrrrr}
\toprule
Timing target & Total used & Fitting & Development & Held-out \\
\midrule
\multicolumn{5}{l}{\emph{Historical training runs}} \\
Custom training loop (s) & 586 & 465 & 121 & -- \\
Trainer-reported runtime (s) & 215 & 170 & 45 & -- \\
\midrule
\multicolumn{5}{l}{\emph{Historical inference runs}} \\
Forward pass (ms) & 643 & 484 & 159 & -- \\
Causal LM forward sum (ms) & 667 & 502 & 165 & -- \\
Encoder--decoder forward sum (ms) & 164 & 123 & 41 & -- \\
Generation call (ms) & 29 & 22 & 7 & -- \\
\midrule
\multicolumn{5}{l}{\emph{Controlled generation configurations}} \\
Generation call (ms) & 64 & 32 & 16 & 16 \\
\bottomrule
\end{tabular}
\vspace{-0.2cm}
\caption{Data used in the reported comparisons. Historical rows count
records; the controlled row counts configurations, each measured in
three separate program launches. Fitting data supply learning and
examples. Historical evaluation uses development records; controlled
generation additionally has a held-out partition. A dash means no
reported held-out evaluation. Trainer runtime is reported by Hugging
Face Trainer; forward sums exclude other generation-loop work.}
\label{tab:population-overview}
\vspace{-0.2cm}
\end{table*}

\subsection{Data collection}
\label{sec:collection}

We consulted MLPerf reference workload scripts~\citep{mattson2020mlperf,reddi2019mlperf}
when developing our data-collection pipeline. We used project-specific
scripts to vary workload configurations and collect runtime measurements
on a Slurm-managed university GPU cluster. All reported measurements
were generated through our own executions, rather than imported from
an existing MLPerf measurement dataset or collected from unrelated
users' jobs. Project generators varied models, GPU types, numerical
precision, batch sizes, and workload sizes, and produced launch
scripts specifying the requested resources and execution settings.

The workload scripts use PyTorch and libraries including Torchvision,
Hugging Face Transformers, and Diffusers, with project-specific
configuration, timing, and logging. Each script records configuration
metadata and measurements in a JSON report.

Dataset builders assembled the JSON reports into CSV records and
deduplicated job identifiers. We interpreted workload configurations
using retained launch arguments, source-defined settings, and report
metadata, and checked timing fields against the corresponding logs.
Appendix~\ref{app:history} documents historical curation, including
outcome-dependent filtering, and accounts for records outside the
evaluated populations.

\subsection{Historical runs}

Inference runs cover 22 models in three groups: text encoders
(BERT, RoBERTa, and DistilBERT; 478 runs;
\citealp{devlin2019bert,liu2019roberta,sanh2019distilbert}),
vision models (convolutional networks and ViT; 329 runs;
\citealp{dosovitskiy2021image}), and causal and encoder--decoder
language models (GPT-2, Llama, Qwen, T5, and BART; 1,077 runs;
\citealp{radford2019language,touvron2023llama,raffel2020exploring,lewis2020bart}).
They ran on four GPU types, A100 (509 runs), A40 (529),
Quadro RTX (451), and T4 (395), in FP32 (827), FP16 (1,047),
or BF16 (10). Forward-pass runs use batch sizes from 1 to 32
with text lengths from 64 to 512 tokens or $224 \times 224$
images. Causal LM and encoder--decoder runs use batch sizes
from 1 to 4, input lengths from 32 to 512, and output limits
from 32 to 256. Generation-call runs use batch size 1, FP16,
input lengths from 128 to 256, and an output limit of 64.

Training runs fall into two groups. Custom training loops
(586 runs) train text encoders, GPT-2 variants, convolutional
and vision-transformer models, and diffusion models in FP16
or FP32, covering 21 models in the fitting partition and
18 in development. Trainer runs (215) fine-tune five
decoder-only models in FP16, and their runtime is reported
by the training framework. A few summarization and
synthetic-video runs appear only in the fitting partition.

Of the 1,041 training-labeled records in the archive, the
fixed comparison uses 801 single-GPU, single-process records
with source-supported timing targets, usable configuration
evidence, and corroborated completion. Appendix~\ref{app:history} describes the excluded runs.

\subsection{Timing targets}
\label{sec:historical-time}

The historical scripts time different parts of execution,
so we keep each as a separate target:
\begin{itemize}[noitemsep, topsep=0pt]
\item \textbf{Forward pass}: one synchronized forward call,
with the model and inputs already on the GPU.

\item \textbf{Causal LM forward sum}: during generation
with a decoder-only model, the total time of the prompt
forward pass and the subsequent cached forward passes,
excluding token selection and other loop overhead.

\item \textbf{Encoder--decoder forward sum}: the total
time of encoder and decoder forward computations during
generation, excluding other loop overhead.

\item \textbf{Generation call}: wall-clock time of a
complete \texttt{generate()} call, including decoding
and token selection.

\item \textbf{Custom training loop}: time spent in the
training loop after the model and data are prepared,
covering data loading, forward and backward passes,
optimizer steps, and logging. Most scripts do not
synchronize the GPU before stopping the timer, so this
is host-side loop time.

\item \textbf{Trainer-reported runtime}: the training time
returned by Hugging Face Trainer, checked against the
run's log.
\end{itemize}

None of these targets includes scheduler queue time. Forward sums
measure selected network computations, not end-to-end generation.

\subsection{Controlled generation collection}
\label{sec:new-panel}

Most historical inference records time forward computations
rather than complete generation calls; the generation-call
target has only seven development records from six
configuration groups. To extend this coverage and assess
measurement variability, we measured a full factorial design
on one A100-PCIE-40GB GPU, varying generation settings under
a common procedure. Two workload models, the decoder-only
Qwen2.5-0.5B \citep{qwen2024qwen25} and the encoder--decoder
T5-small \citep{raffel2020exploring}, are crossed with batch
sizes $\{1,2,4,8\}$, input lengths $\{128,512\}$, output limits
$\{64,128\}$, and greedy or four-beam decoding, giving
$2 \times 4 \times 2 \times 2 \times 2 = 64$ configurations.
All runs use FP32, eager attention, KV caching, and one
returned sequence per input, with TF32 disabled. Generation
may stop early at an end-of-sequence token.

\paragraph{Inputs.}
Each workload model uses a fixed bank of eight synthetic
sequences of 512 token IDs, sampled by hashing from the
model's vocabulary with special tokens excluded. Smaller
batches and shorter inputs use prefixes of this bank.
This keeps input construction independent of measured
runtimes; configurations share input prefixes.

\paragraph{Measurement.}
Each configuration is measured in three separate program
launches. Each launch loads the model anew, runs two warm-up
calls, and then times ten calls, for 1,920 timed calls in
total. The timer encloses only the \texttt{generate()} call,
with GPU synchronization immediately before it starts and
before it stops; model loading, tokenization, and data
transfer are excluded. The target for a configuration is
the mean of its three launch means. Across the 48 fitting
and development configurations, the median coefficient of
variation is 0.27\% within a launch and 3.20\% across launches,
although a few configurations vary by up to 20\%.
We also record how many tokens each call actually generated,
as an outcome rather than an input.
Appendix~\ref{app:collection} lists the software environment
and file hashes.

\subsection{Partitions and evaluation roles}
\label{sec:data-splits}

Fitting records, called the training split in our experiments,
supply regressor training, lookup histories, and five-shot examples.
Development records support method inspection and selection.
Held-out records are scored after methods and predictions are fixed.
Repeated measurements of a configuration remain in the same partition
(Appendix~\ref{app:schemas}).

Historical training uses 635 fitting and 166 development
records. Inference uses 1,131 fitting and 372 development
records: 1,503 of the 1,884-record inventory. The remaining
381 were assigned to a test partition but inspected during
earlier development; we exclude them from these comparisons.
Historical evaluation therefore uses 538 development queries,
not an untouched test set. The seven-query generation-call
comparison is exploratory.

Controlled generation uses 32 fitting, 16 development, and
16 held-out configurations, equally divided between Qwen
and T5 in each partition. Predictions were fixed before
held-out aggregate runtimes were accessed for comparative
scoring. This evaluates combinations of known settings,
not unseen models or hardware. Earlier collection-quality
checks and shared inputs limit its independence; it is
not external confirmation.
We do not pool development and held-out results or different timing targets.

%% file: evaluation.tex
\input{generated_main}

\section{Experimental Setup}
\label{sec:evaluation}

We use the partitions described in Section~\ref{sec:data-splits}.
Within each collection and timing target, methods receive the same
eligible evaluation queries. No predictor was refit on fitting and
development records combined.

\paragraph{Regression and constant baselines.}
We compare ordinary least squares (OLS), Ridge regression,
Random Forest \citep{breiman2001random}, and LightGBM
\citep{ke2017lightgbm}. Features describe \emph{model/task identity},
\emph{workload size} (batch size, input dimensions, and
training-step or output limits), \emph{hardware} (GPU type
and, where applicable, count), and \emph{execution settings}
(precision, optimizer and adapter settings for training,
or decoding and cache settings for controlled generation).
The exact fields vary by target; all four regressors use
the same fields within each target
(Appendix~\ref{app:schemas}).
For historical runs, configuration values are recovered from
job-launch arguments and default settings in the saved workload
scripts. The regressors use no code-analysis features,
measured resource usage, or realized output lengths. Modeled operation counts, memory traffic, and analytical runtime
estimates are additional inputs only in the separate analytical
comparison, not in the main regression baselines.

The \emph{median baseline} predicts the
fitting-set median runtime for every query of a timing target,
providing a reference that ignores workload features.
The four regressors predict log runtime. Numeric features are
standardized and categorical features one-hot encoded using
fitting-set statistics, with an explicit unknown category.
We use fixed hyperparameters (Appendix~\ref{app:settings}),
except for LightGBM on controlled generation: its minimum of
20 examples per leaf prevents splitting with 32 fitting
configurations, so we lowered it to 4 after inspecting development
results. We retain the original setting for reference.

\paragraph{Nearest-neighbor lookup.}
As a history-based reference, \emph{NN lookup}
predicts the median runtime of the closest compatible fitting
records and abstains when none exists; matching rules are specified
in Appendix~\ref{app:schemas}. On historical targets,
\emph{NN lookup + LightGBM} falls back to LightGBM whenever
lookup abstains.

\paragraph{Analytical estimate.}
For FP32 forward passes of transformer encoders and ViT, we compute
a roofline-style estimate from modeled FLOPs, memory traffic, and
nominal GPU peaks. This separate study uses 106 fitting and
42 development records. We compare the estimate with a LightGBM
multiplicative correction, a direct LightGBM predictor, and an
out-of-fold stack of the direct and corrected predictions.
Both fitted models receive the same additional static features,
beyond the primary comparison's inputs.
Appendix~\ref{app:analytical} details the calculation and retained
A100 hardware-label discrepancy.

\paragraph{Predictor LLMs.}
We prompt Qwen3.5-9B \citep{qwen35modelcard},
GPT-OSS-120B \citep{gptossmodelcard}, and
Claude Opus 5.5 \citep{anthropicopus55} with the timing definition
and workload configuration, asking for a number in a fixed JSON
format (Appendix~\ref{app:prompt}). Five-shot prompts also include
five fitting examples with measured runtimes. Historical examples
match the query's timing target. For controlled generation, the
same five configurations (four Qwen and one T5 workload, chosen
by fixed hash ordering) are used for every query and LLM.
Qwen and GPT-OSS use greedy decoding and a 512-token output limit,
with thinking disabled for Qwen and low reasoning effort for
GPT-OSS. Claude uses adaptive thinking at medium effort and a
4,096-token limit including reasoning. Responses must satisfy
the frozen parser's format and unit requirements and yield a
finite positive prediction; failures are retained without
regeneration. Supervision and inference budgets differ:
regressors use all fitting records, whereas five-shot LLMs
receive five examples and bring pretrained knowledge.

\paragraph{Metrics.}
Let $\mathcal{V}$ be the queries with valid predictions, with
measured runtime $y_i>0$ and prediction $\hat{y}_i$.
We report mean absolute error (MAE) in the target's unit and
mean absolute percentage error (MAPE):
\begin{equation}
\begin{split}
\mathrm{MAE}  &= \frac{\sum_{i\in\mathcal V}|\hat y_i-y_i|}{|\mathcal V|},\\
\mathrm{MAPE} &= \frac{100}{|\mathcal V|}\sum_{i\in\mathcal V}\frac{|\hat y_i-y_i|}{y_i}.
\end{split}
\end{equation}
MAE measures absolute timing error; MAPE measures relative error,
giving greater weight to short runtimes for the same absolute
error. Both condition on valid predictions, so coverage
differences can change the rows being averaged. We report
coverage separately and use common valid queries for paired
comparisons. We also report median absolute percentage error
and signed relative error (positive means overprediction).
\emph{Within-factor-two accuracy} is the fraction of all queries
with $y_i/2 \leq \hat{y}_i \leq 2y_i$: for a measured 100\,ms,
predictions from 50 to 200\,ms qualify. Invalid predictions count
as misses. Controlled-generation metrics weight configurations
equally. We compute 95\% confidence intervals by resampling
configuration groups \citep{efron1993introduction}, with
2,000 bootstrap draws for historical data and 10,000 paired
draws for the controlled holdout
(Appendix~\ref{app:settings}).

%% file: generated_main.tex
\begin{table*}[t]
\centering\small
\setlength{\tabcolsep}{2pt}
\renewcommand{\arraystretch}{1.08}
\begin{tabular*}{\textwidth}{@{\extracolsep{\fill}}l@{\hspace{7pt}}rr@{\hspace{7pt}}rr@{\hspace{7pt}}rr@{\hspace{7pt}}rr@{\hspace{7pt}}rr@{\hspace{7pt}}rr@{}}
\toprule
 & \multicolumn{4}{c}{Training duration} & \multicolumn{8}{c}{Inference duration}\\
\cmidrule(lr){2-5}\cmidrule(lr){6-13}
 & \multicolumn{2}{c}{\shortstack{Custom loop\\$n=121$}}
 & \multicolumn{2}{c}{\shortstack{Trainer\\$n=45$}}
 & \multicolumn{2}{c}{\shortstack{Forward pass\\$n=159$}}
 & \multicolumn{2}{c}{\shortstack{Causal LM\\forward sum\\$n=165$}}
 & \multicolumn{2}{c}{\shortstack{Encoder--decoder\\forward sum\\$n=41$}}
 & \multicolumn{2}{c}{\shortstack{Generation call\\$n=7$}}\\
\cmidrule(lr){2-3}\cmidrule(lr){4-5}\cmidrule(lr){6-7}
\cmidrule(lr){8-9}\cmidrule(lr){10-11}\cmidrule(lr){12-13}
Method & MAPE & \shortstack{Within\\factor 2}
 & MAPE & \shortstack{Within\\factor 2}
 & MAPE & \shortstack{Within\\factor 2}
 & MAPE & \shortstack{Within\\factor 2}
 & MAPE & \shortstack{Within\\factor 2}
 & MAPE & \shortstack{Within\\factor 2}\\
\midrule
Median & 149.8 & 43 & 50.5 & 80 & 87.0 & 53 & 150.2 & 43 & 135.0 & 56 & 24.8 & 100\\
OLS & 30.9 & 92 & 27.6 & 98 & 46.6 & 81 & 22.6 & 100 & 21.8 & 100 & 8.7 & 100\\
Ridge & 30.9 & 92 & 27.6 & 98 & 46.9 & 80 & 22.7 & 100 & 21.9 & 100 & \textbf{6.6} & 100\\
Random Forest & 36.4 & 84 & 33.0 & 100 & 34.5 & 89 & 7.6 & 100 & \textbf{7.1} & 100 & \textbf{6.6} & 100\\
LightGBM & 45.7 & 87 & 27.5 & 100 & 29.4 & 92 & \textbf{7.3} & 100 & 7.9 & 100 & 26.7 & 100\\
\midrule
NN lookup & 53.3$^\dagger$ & 54 & 45.7$^\dagger$ & 84 & 62.6 & 74 & -- & -- & -- & -- & 14.0$^\dagger$ & 71\\
\midrule
Qwen 0-shot & 505.1 & 31 & 38.7 & 62 & 116.1 & 45 & 91.7 & 2 & 79.4 & 22 & 70.4 & 43\\
Qwen 5-shot & 372.1 & 21 & 38.0 & 84 & 57.1 & 53 & 136.5 & 45 & 203.8 & 51 & 50.4 & 71\\
GPT-OSS 0-shot & 237.6$^\dagger$ & 30 & 68.9$^\dagger$ & 22 & 67.6 & 43 & 93.9 & 0 & 90.4 & 5 & 88.1 & 0\\
GPT-OSS 5-shot & 329.8 & 26 & 64.4 & 56 & 89.1 & 57 & 153.9 & 48 & 65.6 & 54 & 40.6 & 86\\
Claude 0-shot & \textbf{27.4} & 93 & 41.3 & 80 & 20.9$^\dagger$ & 96 & 20.8 & 96 & 67.4 & 22 & 29.6 & 57\\
Claude 5-shot & 38.6 & 92 & \textbf{19.6} & 98 & \textbf{20.9} & 96 & 14.9 & 100 & 22.4 & 100 & \textbf{6.6} & 100\\
\bottomrule
\end{tabular*}
\vspace{-0.2cm}
\caption{\textbf{Historical development results}
($n$: records). Both metrics are percentages.
MAPE: mean absolute percentage error on valid predictions
(lower is better). Within factor 2: predictions between half
and twice measured runtime (higher is better; failures count
as misses). $^\dagger$: partial coverage; --: no predictions.
Bold: lowest full-coverage MAPE, including rounding ties.
Trainer: Hugging Face Trainer. MAE and coverage:
Appendix~\ref{app:full-results}.}
\label{tab:overview}\end{table*}

\begin{table}[!t]
\centering\small
\setlength{\tabcolsep}{3pt}
\begin{tabular}{lrrrr}
\toprule
Method & \shortstack{MAE\\(ms)} & \shortstack{MAPE\\(\%)} & \shortstack{Signed\\RE (\%)} & \shortstack{Within\\factor 2\\(\%)}\\
\midrule
Median & 667.2 & 64.3 & 25.3 & 50.0\\
OLS & 80.8 & 6.0 & 1.6 & 100.0\\
Ridge & 83.3 & 6.2 & 1.6 & 100.0\\
Random Forest & \textbf{59.9} & 5.0 & 1.5 & 100.0\\
LightGBM & 64.6 & \textbf{4.6} & 3.1 & 100.0\\
LightGBM (leaf 20) & 667.2 & 63.6 & 24.0 & 50.0\\
NN lookup & 99.0 & 9.2 & $-$1.9 & 100.0\\
\midrule
Qwen 0-shot & 1275.2 & 98.6 & $-$98.6 & 0.0\\
Qwen 5-shot & 566.1 & 55.8 & 20.5 & 68.8\\
GPT-OSS 0-shot & 1166.7 & 89.9 & $-$89.9 & 0.0\\
GPT-OSS 5-shot & 593.4 & 57.2 & 33.2 & 68.8\\
Claude 0-shot & 271.4 & 19.0 & 8.8 & 100.0\\
Claude 5-shot & 88.8 & 8.2 & $-$4.6 & 100.0\\
\bottomrule
\end{tabular}
\vspace{-0.2cm}
\caption{\textbf{Generation-call duration on the controlled
holdout} (16 configurations; full coverage).
MAE: mean absolute error (ms/batch); MAPE: mean absolute
percentage error; RE: signed relative error (positive:
overprediction). Within factor 2 accepts 50--200\,ms for
a measured 100\,ms. All except MAE are percentages.
LightGBM uses minimum leaf size 4 unless specified.
Bold: lowest MAE/MAPE.}
\label{tab:locked}
\end{table}

%% file: results.tex
\section{Results}
\label{sec:results}

Table~\ref{tab:overview} summarizes historical development MAPE
and within-factor-two accuracy, and Table~\ref{tab:locked}
reports results on the 16 controlled holdout configurations.

\paragraph{Zero-shot accuracy varies across LLMs and targets.}
On the controlled holdout, Qwen3.5-9B and GPT-OSS-120B underestimate
latency by 99\% and 90\% on average, and none of their predictions
falls within a factor of two of the measured value
(Table~\ref{tab:locked}). Accuracy is also low on historical
causal LM forward sums, where only 2\% and 0\% of their predictions
fall within a factor of two. Claude Opus 5.5 is far more accurate,
with 19.0\% MAPE on the controlled holdout, but its zero-shot
accuracy depends strongly on the target. It has the lowest
observed MAPE on custom training loops (27.4\% MAPE), yet on
encoder--decoder forward sums only 22\% of its predictions fall
within a factor of two, and its MAE (835.0~ms) is more than ten
times that of LightGBM (79.6~ms).

\paragraph{Five examples help, but unevenly.}
On the controlled holdout, five examples lower every LLM's MAE
and MAPE; Claude's MAE falls from 271.4 to 88.8~ms.
On historical targets, Claude's MAE and MAPE improve on four
of six targets. Its MAE falls by 65\% for training time reported
by Hugging Face Trainer (238.7 to 83.6~s), but rises for custom
training loops (157.2 to 172.8~s). Forward-pass MAE also rises
slightly, from 5.45 to 5.51~ms on the same 158 valid queries;
five-shot answers all 159 queries, whereas zero-shot has one
invalid response. For custom loops, absolute error increases on
69 queries, decreases on 49, and ties on three. Predictions
increase on 97 of 121 queries, but the audit does not establish
why. With five examples, the open-weight LLMs still have
56--57\% MAPE on the controlled holdout, and their factor-two
accuracy on custom loops drops (Table~\ref{tab:overview}).

\paragraph{Simple regressors are hard to beat.}
Regressors trained on \bench{} provide competitive predictions,
but their ranking relative to LLMs depends on the target and
metric. By MAE, linear models are the best regressors on custom
training loops, Random Forest on Trainer-reported runtimes,
and LightGBM on the two forward sums, where its MAE is 39\%
(causal LM) and 72\% (encoder--decoder) lower than that of
Claude five-shot. On the controlled holdout, Random Forest has
the lowest MAE (59.9~ms) and LightGBM the lowest MAPE (4.6\%).
The 95\% interval for Claude five-shot minus Random Forest MAE
spans zero ($-21.3$ to $73.4$~ms), so it does not establish an
ordering. The corresponding MAPE interval against LightGBM
is positive (0.1 to 7.3 percentage points), but these pairwise
intervals are unadjusted for multiple comparisons
(Appendix~\ref{app:pairs}). 

\paragraph{Errors depend on the workload model.}
On the controlled holdout, Claude five-shot has similar observed
MAPE to Random Forest on Qwen2.5-0.5B configurations
(5.1\% vs.\ 5.0\% MAPE), but on T5-small configurations its MAPE
is almost three times that of LightGBM (11.4\% vs.\ 4.1\%).
With five examples, the open-weight LLMs overestimate T5 latency
by more than 50\% on average. Four of the five examples are
Qwen workloads, but this fixed example set does not establish
whether its composition explains the difference. Each slice
contains only eight configurations, so this analysis is
exploratory.

\paragraph{Lookup and analytical estimates.}
NN lookup requires compatible fitting history to return a
prediction. It answers all forward-pass queries, 80 of 121
custom-loop queries, and 43 of 45 Trainer queries, but none
of the forward-sum queries, for which the compatibility
information it needs is not recorded. Falling back to LightGBM
gives full coverage, but on custom loops and Trainer runtimes
the combination has higher MAE than LightGBM alone
(286.5 vs.\ 236.9~s and 201.3 vs.\ 162.1~s).
In the FP32 forward-pass study (42 development runs;
Table~\ref{tab:fp32}), the analytical estimate alone
underestimates runtime by 56\% on average, but learning a
correction on top of it reduces MAE from 22.8 to 2.75~ms,
well below the 11.9~ms of a direct LightGBM model given
the same features. Stacking the direct and corrected
predictions gives the lowest MAPE (12.1\%) in this study.
These results retain the nominal hardware assumptions
described in Appendix~\ref{app:analytical}.

%% file: appendix.tex
\raggedbottom
\makeatletter
\setlength{\@dblfptop}{0pt}
\setlength{\@dblfpsep}{12pt}
\setlength{\@dblfpbot}{0pt plus 1fil}
\makeatother
\section{Historical Data Details}
\label{app:history}

\paragraph{Accounting.}
The retained archive contains 1,041 training-labeled records
and 1,884 inference records. Our experiments use 801 training
records and 1,503 inference records from the fitting and
development partitions. The remaining 381 inference records
belong to the original test partition: 164 forward-pass,
169 causal LM forward-sum, 41 encoder--decoder forward-sum,
and seven generation-call records. This partition was examined
during earlier development and is not evaluated here.
Including these records, the four inference targets contain
807, 836, 205, and 36 records, respectively.

The 240 training-labeled records outside the fixed comparison
comprise 118 records with supported timing measurements and
122 records with unresolved configuration or completion.
The 118 comprise 32 custom-loop records, 28 DataParallel-loop
records, and 58 records whose Trainer runtime was reported by
the primary process (rank zero). Supported timing alone does
not establish eligibility: input reconstruction for 117 of
these records depends on additional assumptions, while one
has an unresolved input issue. None was added to the fixed
comparison. Of the other 122 records, 111 contain an end
marker but have unresolved workload configurations, and
11 have unconfirmed application completion. These excluded
records should therefore not all be described as failed runs.

Earlier exports contain 1,337 distinct training-labeled job
IDs, of which 1,041 are represented in the retained archive.
The 296 absent IDs are not additional verified failures.
Historical curation included outcome-dependent filtering and
protocol selection, so the retained records are not an
unfiltered sample of all attempted workloads.

\paragraph{Earlier outcome-dependent filtering.}
Retained cleaning code groups runs by model, GPU, precision,
batch size, epochs, batches per epoch, parallelism, GPU count,
dataset, and sequence length. For groups with at least two
positive runtimes, it rejects deviations from the median
exceeding both 40\% of the median and an additional threshold:
2.5 times the median absolute deviation (MAD) when MAD is
at least one second, or 50\% of the median otherwise.
Later passes reject positive data-loading/runtime ratios
above 0.8 and actual/analytical runtime ratios above 100.

An earlier curation report records 11 outlier exclusions,
185 loading-ratio exclusions, and no ratio-above-100 exclusions
in a revision from 1,070 to 874 records. These counts describe
that earlier version; they cannot be subtracted from the
current archive or treated as the explanation for all
296 absent IDs. Protocol selection also considered observed
variability and prediction performance. The historical
collection is therefore outcome-selected. We preserve the
fixed evaluated population without rerunning these filters
or adding recovered records.

\paragraph{Inputs and hardware.}
Historical text-encoder inference runs use synthetic token IDs
and attention masks of shape $[B,S]$, where $B$ is batch size
and $S$ is sequence length. Vision inference runs use tensors
of shape $[B,3,224,224]$. GPU types---A100, A40, T4, and
Quadro RTX---come from the run metadata.

Source hashes identify the retained scripts used to reconstruct
configurations, rather than necessarily the exact revisions
executed historically. Packaged environment metadata records
Python 3.11 and PyTorch 2.5.1+cu121, but does not establish
software versions for every historical run.
Appendix~\ref{app:collection} documents the controlled
collection's environment separately.

\paragraph{Timing details.}
Forward-pass scripts time one synchronized call with the
model and input tensors already on the GPU. Forward sums
exclude token selection and other generation-loop work;
they measure neither time to first token nor complete
generation latency. Historical generation-call runs include
beam search and sampling, but their realized output lengths
and individual call timings were not recorded.

Custom-loop totals start after model, data, and optimizer
setup. They include the first training step, data loading,
forward and backward passes, parameter updates, and in-loop
logging. Most scripts read the scalar loss but do not explicitly
synchronize the GPU immediately before stopping the timer;
the video-training scripts do synchronize explicitly.
These targets retain the timing behavior of their source
scripts. Trainer targets use the runtime returned by
Hugging Face Trainer, checked against the printed log.

\paragraph{Training configuration fields.}
Epoch limits, per-epoch update caps, and total update limits
remain distinct. Gradient accumulation is recorded separately
from batch size. For DataParallel records in the archive,
the configured batch is not multiplied by the number of GPUs;
these records are outside the reported training comparison.

The 30 input fields describe the model, dataset, task,
hardware, execution settings, workload extent, and optimization
configuration. Appendix~\ref{app:schemas} lists the fields and
their encoding. When reconstruction depends on a library
default, we retain the corresponding assumption or unknown
value rather than substituting a default from the current
environment. Four summarization and four synthetic-video
records occur only in the fitting partition.

\paragraph{Grouping.}
Within each inference timing target, two records are linked
when all configuration fields known for both agree.
Connected components of these links define configuration
groups. Missing fields can therefore join records whose
workloads cannot be distinguished from the available evidence;
this conservative rule keeps possible repetitions together.

The fitting/development group counts are 478/159 for forward
passes, 494/164 for causal LM forward sums, 123/41 for
encoder--decoder forward sums, and 18/6 for generation calls.
Training records are grouped by a hash of all 30 input fields,
including the timing procedure, giving 447/112 groups for
custom loops and 144/37 for Trainer-reported runtime.
These are configuration-group counts, not record counts.
Grouping uses configuration information, not measured outcomes.
Appendix~\ref{app:schemas} specifies the partition-assignment rules.

\section{Predictor Settings}
\label{app:settings}

\paragraph{Regression and constant baselines.}
The median baseline returns the median runtime of the fitting
records for every query. OLS uses minimum-norm least squares,
and Ridge uses penalty 1 with an unpenalized intercept.
Numeric preprocessing uses fitting-partition medians for
imputation and fitting-partition means and population standard
deviations for standardization, with a standard-deviation
floor of $10^{-9}$. Categorical features use full one-hot
encoding with an unknown category. Tree models receive the
same features without the intercept column. In the controlled
collection, invalid required inputs are rejected rather than
imputed.

Random Forest \citep{pedregosa2011scikit} uses 200 trees,
maximum depth 10, minimum leaf size 2, all features at each
split, bootstrap sampling, and seed 0. LightGBM uses 200 rounds,
learning rate 0.05, 15 leaves, maximum depth 5, maximum bin 255,
L1 penalty 0, L2 penalty 1, full feature and bagging fractions,
no early stopping, deterministic column-wise CPU training,
and seed 0.

On the controlled collection, the default minimum of 20
examples per leaf prevents any split with only 32 fitting
configurations. After inspecting development results, we
reduced this minimum to 4. This variant changes only the
minimum leaf size and produces 200 trees with 1,230 splits;
we made no other changes and ran no further search. We also
report the default setting. Ridge and LightGBM clip log
predictions to $[-30,30]$ before exponentiation; OLS and
Random Forest do not apply this clipping.

\paragraph{NN lookup.}
NN lookup computes Euclidean distances in the standardized
and one-hot encoded feature space constructed from fitting
records. It searches compatible records with the same model,
GPU, and precision and returns the median runtime of the
nearest records, retaining ties under the implementation's
tie rule. Historical lookup additionally requires matching
task, strategy, and extent fields.
Appendix~\ref{app:schemas} specifies the target-specific
compatibility and tie rules. Lookup abstains when no compatible
record exists. Its coverage is part of the result, since errors
on answered queries do not describe performance on all queries.

\paragraph{Uncertainty.}
Historical bootstrap intervals use 2,000 group-level draws
with seed 0. Controlled intervals use 10,000 paired draws of
whole configurations with seed 20261003 and the 2.5th and
97.5th percentiles. Paired comparisons resample the same
configurations for both methods. We report intervals only
when at least ten groups are available, following our reporting
threshold, and do not adjust them for multiple comparisons.
These intervals describe variation across the sampled
configuration groups; they do not capture uncertainty from
new hardware, different model versions, or repeated predictor
training and LLM execution.

\paragraph{Predictor LLMs.}
We use Qwen/Qwen3.5-9B, openai/gpt-oss-120b,
and Claude Opus 5.5 as runtime predictors.

Qwen runs in BF16 with Transformers 5.17.0
\citep{wolf2020transformers} and PyTorch 2.5.1+cu121
\citep{paszke2019pytorch}, using SDPA attention, thinking
disabled, greedy decoding, seed 0, an 8,192-token context,
and a 512-token output limit. GPT-OSS runs with MXFP4 weights
and BF16 compute on vLLM 0.13.0 \citep{kwon2023efficient}
with tensor parallelism 2, PyTorch 2.9, CUDA 12.9, and
Transformers 4.57.3, using low reasoning effort, greedy
decoding, seed 0, an 8,192-token context, a 512-token output
limit, eager execution, and prefix caching disabled.

Claude is called through the API as
\texttt{claude-opus-5-5} with adaptive thinking
(not displayed), medium effort, and a 4,096-token output
limit that includes reasoning. No tools or search are enabled
for any LLM. The reasoning settings and output budgets
therefore differ across predictors.

\paragraph{Demonstrations and parsing.}
For the controlled collection, the five demonstrations are
fitting configurations C001, C018, C024, C030, and C047,
in that order, selected by a fixed hash ranking that does
not depend on prediction errors. Four are Qwen workloads
and one is a T5 workload. The same five demonstrations are
used for every controlled query and every predictor LLM.

For historical targets, the selector excludes the query's
group, takes the lexicographically first run in each fitting
group, ranks groups by the SHA-256 hash of
\texttt{["configuration-llm-v1", "examples", 0, procedure, group]},
and uses the first five.

Each predictor is issued 16 zero-shot and 16 five-shot
requests for each controlled evaluation partition
(development and holdout), and 538 historical development
queries per setting. A response must satisfy the frozen
format and extraction rules and contain a finite positive
number in the required unit. Invalid responses count as
failures and are never repaired or regenerated.

\section{Controlled Collection Details}
\label{app:collection}

\paragraph{Environment.}
Measurements use Python 3.11.14, PyTorch 2.5.1+cu121
(CUDA 12.1), Transformers 5.2.0, and tokenizers 0.22.2.
The scientific manifest identifies the model weights,
tokenizers, configuration files, and input banks by
SHA-256 hashes.

\paragraph{Input construction.}
We generate synthetic inputs from each model's sorted
list of valid, non-special vocabulary IDs. Token selection
is deterministic, using a SHA-256 hash of the model name,
seed 20261002, example index, and token position.
Token bounds, input-bank hashes, attention masks, and
model files are checked before execution.

Stopping at the end-of-sequence (EOS) token is enabled,
the minimum number of new tokens is zero, and one sequence
is returned per input. Beam search uses length penalty 1
and early stopping. Consequently, the requested output
limit need not equal the number of tokens returned.

\paragraph{Measurement.}
The collector runs in inference mode with gradients disabled.
Each generation call builds its own cache; caches are not
reused across calls. A monotonic high-resolution host clock
times only the \texttt{generate()} call, with GPU
synchronization immediately before and after it.
Model loading, input preparation, and warm-up calls are
outside the timed interval.

Startup timings and peak memory are recorded separately
as diagnostics and are not predictor inputs. A preliminary pilot using
natural-text inputs tested the collection pipeline; its
measurements are excluded from the benchmark targets.

\paragraph{Generated-token counts.}
We count returned tokens after removing the input or
decoder-start prefix, including the first end-of-sequence
token and excluding subsequent padding. These counts are
recorded as outcomes and are never supplied to predictors.

\paragraph{Evaluation order.}
Split assignments were fixed before collection.
Methods, prompts, and parsers were finalized after
development evaluation. All holdout predictions were saved
before aggregate runtime labels were accessed for comparative
scoring. Earlier collection-quality access is described in
Section~\ref{sec:data-splits}. No predictor setting was changed
after holdout scoring.
\section{Compute and Release}
\label{app:resources}

\paragraph{Compute.}
Collecting the controlled dataset required approximately
4.4 GPU-hours, with another 1.3 GPU-hours for preliminary
pilots. Running the open-weight predictor LLMs required
approximately 0.4 GPU-hours on the controlled development
partition and 0.4 GPU-hours on the holdout, using one A100
for Qwen and two A40s for GPT-OSS. These figures cover the
controlled study, not historical data collection.
Regression fitting and scoring took a few CPU-minutes;
Claude predictions used the provider's API.

\paragraph{Release status.}
Reviewer data and code packages have been prepared and
verified for offline reproduction of the reported scores.
Anonymous hosting and redistribution permissions remain
pending. The packages include reference collection and
prediction code, but end-to-end regeneration has not
been verified.

\section{Complete Results}
\label{app:full-results}

The following tables report results for all evaluated methods
and timing targets. MAE is measured in the target's time unit;
MAPE and median absolute percentage error (Med.\ APE) are
percentages. Signed relative error (RE) is positive for
overprediction and negative for underprediction.
Coverage is the number of valid predictions divided by
the number of eligible queries.

Error metrics use valid predictions only. Within-factor-2
accuracy is the percentage of all eligible queries whose
predictions lie between half and twice the measured runtime;
invalid responses count as misses. Controlled results weight
each configuration equally.

Confidence intervals follow Appendix~\ref{app:settings}.
The historical generation-call evaluation has six configuration
groups, and each controlled workload-model slice has eight;
both fall below our ten-group threshold for reporting intervals.
The workload-model slices are exploratory.

\input{generated_historical}
\input{generated_test_full}
\input{generated_synthetic}

\clearpage
\section{Analytical Forward-Pass Study}
\label{app:analytical}

We evaluate analytical estimates on 106 fitting and
42 development records of FP32 forward passes from
transformer encoders and ViT. The estimate uses modeled
computation, memory traffic, and nominal GPU peaks without
profiling the query workload.

We compare the analytical estimate, a learned multiplicative
correction, a direct LightGBM predictor, and a stack of the
two learned predictors. The direct and correction models
receive identical inputs, including the analytical estimate,
but learn different targets: log runtime and log correction,
respectively. All fitting uses the fitting partition;
stacking uses three internal folds.

\input{generated_analytical}
\input{analytical_contract}

\clearpage
\input{generated_pairs}
\clearpage

\section{Prompt}
\label{app:prompt}

Below is the exact system instruction used for controlled
generation queries; only line breaks have been added.
Historical prompts specify their corresponding timing
target and unit.

\begin{quote}\small
Predict resident-input generation-call wall latency in milliseconds per batched invocation. The model and input tensors are already resident on the GPU. The host monotonic timer surrounds the full model.generate call with GPU synchronization immediately before start and before stop. Exclude loading, tokenization, input transfer, startup, warmup, validation and serialization. The target is the equally weighted mean of three fresh-process launch means, each over ten measured calls after two warmups. This is a controlled synthetic token-ID input study with shared nested input banks, not natural prompts. Natural EOS remains enabled. Requested output cap is a limit, not an observed generated count. Returned beam sequences are not all internal beam work. Use only the supplied declarations and any provided fitting demonstrations; do not request tools or measurements. Return exactly one JSON object with keys prediction\_ms (finite positive number) and unit (exactly "ms\_per\_batch"). No explanation or markdown.
\end{quote}

The following development query is shown with formatting
added for readability. It contains neither a record identifier
nor the measured runtime. Five-shot requests include the
five fitting demonstrations listed in
Appendix~\ref{app:settings} as preceding user--assistant
pairs; zero-shot requests contain no demonstrations.

\begin{small}
\begin{verbatim}
{
  "attention_backend": "eager",
  "batch_size": 1,
  "declared_gpu_sku":
    "NVIDIA A100-PCIE-40GB",
  "do_sample": false,
  "input_length": 128,
  "model": "qwen2.5-0.5b",
  "num_beams": 4,
  "num_return_sequences": 1,
  "precision": "float32",
  "requested_gpu_count": 1,
  "requested_output_cap": 64,
  "seed": 20261002,
  "task": "causal_generation",
  "tf32": false,
  "use_cache": true
}
\end{verbatim}
\end{small}

\input{technical_specifications}

%% file: generated_historical.tex
\begin{table*}[tp]
\centering\small
\begin{tabular}{lrrrrrr}
\toprule
Method & Coverage & MAE (s) & MAPE (\%) & Med. APE & Signed RE & \shortstack{Within\\factor 2 (\%)}\\
\midrule
Median & 121/121 & 486.64 & 149.80 & 69.84 & 95.85 & 42.98\\
OLS & 121/121 & 163.39 & 30.86 & 21.54 & 13.81 & 91.74\\
Ridge & 121/121 & 161.94 & 30.94 & 22.85 & 13.69 & 91.74\\
Random Forest & 121/121 & 221.37 & 36.43 & 23.15 & 11.56 & 84.30\\
LightGBM (leaf 20) & 121/121 & 236.91 & 45.65 & 23.23 & 27.48 & 86.78\\
NN lookup & 80/121 & 274.53 & 53.29 & 14.95 & 24.29 & 53.72\\
NN lookup + LightGBM & 121/121 & 286.49 & 56.45 & 21.82 & 31.33 & 80.17\\
Qwen zero-shot & 121/121 & 1294.24 & 505.13 & 76.88 & 431.26 & 30.58\\
Qwen five-shot & 121/121 & 735.61 & 372.06 & 91.14 & 292.17 & 20.66\\
GPT-OSS zero-shot & 108/121 & 589.77 & 237.64 & 78.47 & 143.65 & 29.75\\
GPT-OSS five-shot & 121/121 & 911.31 & 329.76 & 88.42 & 259.78 & 25.62\\
Claude zero-shot & 121/121 & 157.19 & 27.43 & 21.11 & 1.46 & 93.39\\
Claude five-shot & 121/121 & 172.80 & 38.64 & 26.26 & 29.48 & 91.74\\
\bottomrule
\end{tabular}
\caption{Custom training loop, historical dev split. Errors are computed over valid predictions, and within-2$\times$ accuracy counts failures as misses. Median APE and signed RE are percentages.}
\label{tab:hist-manual_training_loop_s}
\end{table*}

\begin{table*}[tp]
\centering\small
\begin{tabular}{lrrrrrr}
\toprule
Method & Coverage & MAE (s) & MAPE (\%) & Med. APE & Signed RE & \shortstack{Within\\factor 2 (\%)}\\
\midrule
Median & 45/45 & 336.62 & 50.54 & 47.65 & 13.43 & 80.00\\
OLS & 45/45 & 163.32 & 27.56 & 23.95 & 14.14 & 97.78\\
Ridge & 45/45 & 164.01 & 27.61 & 24.50 & 14.19 & 97.78\\
Random Forest & 45/45 & 136.81 & 33.00 & 28.13 & 18.10 & 100.00\\
LightGBM (leaf 20) & 45/45 & 162.12 & 27.52 & 25.85 & 10.64 & 100.00\\
NN lookup & 43/45 & 210.49 & 45.74 & 37.88 & 14.96 & 84.44\\
NN lookup + LightGBM & 45/45 & 201.26 & 43.78 & 37.18 & 14.22 & 88.89\\
Qwen zero-shot & 45/45 & 316.45 & 38.65 & 36.79 & -35.94 & 62.22\\
Qwen five-shot & 45/45 & 219.86 & 38.01 & 34.06 & -4.15 & 84.44\\
GPT-OSS zero-shot & 43/45 & 400.59 & 68.86 & 77.82 & -61.15 & 22.22\\
GPT-OSS five-shot & 45/45 & 225.16 & 64.43 & 54.05 & 9.04 & 55.56\\
Claude zero-shot & 45/45 & 238.71 & 41.34 & 35.13 & 7.57 & 80.00\\
Claude five-shot & 45/45 & 83.63 & 19.64 & 19.05 & -0.93 & 97.78\\
\bottomrule
\end{tabular}
\caption{Trainer-reported runtime, historical dev split. Errors are computed over valid predictions, and within-2$\times$ accuracy counts failures as misses. Median APE and signed RE are percentages.}
\label{tab:hist-trainer_reported_runtime_s}
\end{table*}

\begin{table*}[tp]
\centering\small
\begin{tabular}{lrrrrrr}
\toprule
Method & Coverage & MAE (ms) & MAPE (\%) & Med. APE & Signed RE & \shortstack{Within\\factor 2 (\%)}\\
\midrule
Median & 159/159 & 20.11 & 87.01 & 70.29 & 30.78 & 52.83\\
OLS & 159/159 & 10.64 & 46.56 & 34.95 & 13.95 & 81.13\\
Ridge & 159/159 & 10.71 & 46.86 & 35.76 & 14.23 & 79.87\\
Random Forest & 159/159 & 9.89 & 34.54 & 25.05 & 5.95 & 89.31\\
LightGBM (leaf 20) & 159/159 & 5.63 & 29.44 & 16.41 & 9.95 & 92.45\\
NN lookup & 159/159 & 15.30 & 62.61 & 45.61 & 14.15 & 73.58\\
NN lookup + LightGBM & 159/159 & 15.30 & 62.61 & 45.61 & 14.15 & 73.58\\
Qwen zero-shot & 159/159 & 18.69 & 116.06 & 78.68 & 80.53 & 44.65\\
Qwen five-shot & 159/159 & 18.13 & 57.13 & 51.15 & -18.74 & 53.46\\
GPT-OSS zero-shot & 159/159 & 19.65 & 67.57 & 64.29 & -23.98 & 42.77\\
GPT-OSS five-shot & 159/159 & 16.87 & 89.07 & 56.21 & 53.35 & 57.23\\
Claude zero-shot & 158/159 & 5.45 & 20.87 & 18.76 & -5.99 & 96.23\\
Claude five-shot & 159/159 & 5.48 & 20.92 & 15.02 & -2.01 & 96.23\\
\bottomrule
\end{tabular}
\caption{Forward pass, historical dev split. Errors are computed over valid predictions, and within-2$\times$ accuracy counts failures as misses. Median APE and signed RE are percentages.}
\label{tab:hist-resident_forward_latency_ms}
\end{table*}

\begin{table*}[tp]
\centering\small
\begin{tabular}{lrrrrrr}
\toprule
Method & Coverage & MAE (ms) & MAPE (\%) & Med. APE & Signed RE & \shortstack{Within\\factor 2 (\%)}\\
\midrule
Median & 165/165 & 1482.99 & 150.23 & 72.40 & 109.17 & 43.03\\
OLS & 165/165 & 419.85 & 22.59 & 23.01 & 6.22 & 100.00\\
Ridge & 165/165 & 415.57 & 22.65 & 23.03 & 6.30 & 100.00\\
Random Forest & 165/165 & 191.36 & 7.56 & 5.07 & 0.90 & 100.00\\
LightGBM (leaf 20) & 165/165 & 186.04 & 7.33 & 5.61 & 0.32 & 100.00\\
NN lookup & 0/165 & -- & -- & -- & -- & 0.00\\
NN lookup + LightGBM & 165/165 & 186.04 & 7.33 & 5.61 & 0.32 & 100.00\\
Qwen zero-shot & 165/165 & 2037.52 & 91.75 & 95.79 & -91.75 & 1.82\\
Qwen five-shot & 165/165 & 1609.74 & 136.53 & 64.94 & 82.46 & 45.45\\
GPT-OSS zero-shot & 165/165 & 2059.52 & 93.89 & 96.70 & -93.89 & 0.00\\
GPT-OSS five-shot & 165/165 & 1879.90 & 153.89 & 68.29 & 118.23 & 48.48\\
Claude zero-shot & 165/165 & 518.29 & 20.83 & 18.46 & -14.99 & 95.76\\
Claude five-shot & 165/165 & 303.47 & 14.87 & 12.97 & 4.62 & 100.00\\
\bottomrule
\end{tabular}
\caption{Causal LM forward sum, historical dev split. NN lookup abstains on every query, so NN lookup + LightGBM equals LightGBM. Errors are computed over valid predictions, and within-2$\times$ accuracy counts failures as misses. Median APE and signed RE are percentages.}
\label{tab:hist-causal_sum_forward_latency_ms}
\end{table*}

\begin{table*}[tp]
\centering\small
\begin{tabular}{lrrrrrr}
\toprule
Method & Coverage & MAE (ms) & MAPE (\%) & Med. APE & Signed RE & \shortstack{Within\\factor 2 (\%)}\\
\midrule
Median & 41/41 & 629.97 & 134.98 & 50.19 & 103.37 & 56.10\\
OLS & 41/41 & 193.12 & 21.76 & 23.98 & 3.83 & 100.00\\
Ridge & 41/41 & 187.87 & 21.86 & 23.63 & 4.11 & 100.00\\
Random Forest & 41/41 & 85.86 & 7.12 & 6.29 & 2.15 & 100.00\\
LightGBM (leaf 20) & 41/41 & 79.60 & 7.94 & 7.17 & 1.60 & 100.00\\
NN lookup & 0/41 & -- & -- & -- & -- & 0.00\\
NN lookup + LightGBM & 41/41 & 79.60 & 7.94 & 7.17 & 1.60 & 100.00\\
Qwen zero-shot & 41/41 & 1029.94 & 79.39 & 93.45 & -79.39 & 21.95\\
Qwen five-shot & 41/41 & 835.69 & 203.82 & 59.39 & 166.81 & 51.22\\
GPT-OSS zero-shot & 41/41 & 1072.86 & 90.37 & 95.56 & -90.37 & 4.88\\
GPT-OSS five-shot & 41/41 & 597.81 & 65.62 & 57.65 & 15.44 & 53.66\\
Claude zero-shot & 41/41 & 835.03 & 67.42 & 67.08 & -67.42 & 21.95\\
Claude five-shot & 41/41 & 279.50 & 22.41 & 25.01 & -15.96 & 100.00\\
\bottomrule
\end{tabular}
\caption{Encoder--decoder forward sum, historical dev split. NN lookup abstains on every query, so NN lookup + LightGBM equals LightGBM. Errors are computed over valid predictions, and within-2$\times$ accuracy counts failures as misses. Median APE and signed RE are percentages.}
\label{tab:hist-seq2seq_sum_forward_latency_ms}
\end{table*}

\begin{table*}[tp]
\centering\small
\begin{tabular}{lrrrrrr}
\toprule
Method & Coverage & MAE (ms) & MAPE (\%) & Med. APE & Signed RE & \shortstack{Within\\factor 2 (\%)}\\
\midrule
Median & 7/7 & 203.54 & 24.82 & 13.32 & 13.74 & 100.00\\
OLS & 7/7 & 103.95 & 8.73 & 7.84 & 8.06 & 100.00\\
Ridge & 7/7 & 68.23 & 6.62 & 6.54 & 4.18 & 100.00\\
Random Forest & 7/7 & 72.25 & 6.59 & 6.19 & 4.99 & 100.00\\
LightGBM (leaf 20) & 7/7 & 255.28 & 26.73 & 22.52 & 1.66 & 100.00\\
NN lookup & 5/7 & 156.89 & 13.96 & 5.02 & 11.76 & 71.43\\
NN lookup + LightGBM & 7/7 & 205.65 & 16.92 & 22.52 & 1.45 & 100.00\\
Qwen zero-shot & 7/7 & 753.89 & 70.37 & 81.61 & -20.09 & 42.86\\
Qwen five-shot & 7/7 & 507.68 & 50.42 & 68.57 & -2.36 & 71.43\\
GPT-OSS zero-shot & 7/7 & 962.74 & 88.07 & 86.89 & -88.07 & 0.00\\
GPT-OSS five-shot & 7/7 & 371.69 & 40.57 & 35.57 & 9.62 & 85.71\\
Claude zero-shot & 7/7 & 341.47 & 29.57 & 5.37 & -24.61 & 57.14\\
Claude five-shot & 7/7 & 61.17 & 6.65 & 3.37 & 5.35 & 100.00\\
\bottomrule
\end{tabular}
\caption{Generation call, historical dev split (seven runs from six configuration groups; exploratory). Errors are computed over valid predictions, and within-2$\times$ accuracy counts failures as misses. Median APE and signed RE are percentages.}
\label{tab:hist-generate_call_latency_ms}
\end{table*}

%% file: generated_test_full.tex
\begin{table*}[tp]
\centering\small
\begin{tabular}{lrrrrrr}
\toprule
Method & Coverage & MAE (ms) & MAPE (\%) & Med. APE & Signed RE & \shortstack{Within\\factor 2 (\%)}\\
\midrule
Median & 16/16 & 667.24 & 64.29 & 53.90 & 25.32 & 50.00\\
OLS & 16/16 & 80.84 & 5.96 & 4.63 & 1.59 & 100.00\\
Ridge & 16/16 & 83.26 & 6.20 & 4.49 & 1.62 & 100.00\\
Random Forest & 16/16 & 59.91 & 4.97 & 4.09 & 1.48 & 100.00\\
LightGBM & 16/16 & 64.63 & 4.61 & 4.17 & 3.09 & 100.00\\
LightGBM (leaf 20) & 16/16 & 667.24 & 63.60 & 53.33 & 23.98 & 50.00\\
NN lookup & 16/16 & 99.05 & 9.25 & 7.50 & $-$1.86 & 100.00\\
Qwen 0-shot & 16/16 & 1275.17 & 98.64 & 98.99 & $-$98.64 & 0.00\\
Qwen 5-shot & 16/16 & 566.08 & 55.83 & 38.68 & 20.48 & 68.75\\
GPT-OSS 0-shot & 16/16 & 1166.74 & 89.89 & 91.97 & $-$89.89 & 0.00\\
GPT-OSS 5-shot & 16/16 & 593.44 & 57.21 & 45.26 & 33.17 & 68.75\\
Claude 0-shot & 16/16 & 271.40 & 18.97 & 17.99 & 8.80 & 100.00\\
Claude 5-shot & 16/16 & 88.83 & 8.22 & 7.63 & $-$4.60 & 100.00\\
\bottomrule
\end{tabular}
\caption{Controlled holdout, all metrics (16 configurations, milliseconds per batch). LightGBM (leaf 4) is the setting used in the main results. Median APE and signed RE are percentages.}
\label{tab:test-full}
\end{table*}

%% file: generated_synthetic.tex
\begin{table*}[tp]
\centering\small
\begin{tabular}{lrrrrrr}
\toprule
Method & Coverage & MAE (ms) & MAPE (\%) & Med. APE & Signed RE & \shortstack{Within\\factor 2 (\%)}\\
\midrule
Median & 16/16 & 659.16 & 66.13 & 46.68 & 28.41 & 50.00\\
OLS & 16/16 & 103.66 & 7.94 & 7.24 & 3.20 & 100.00\\
Ridge & 16/16 & 95.64 & 8.00 & 7.55 & 3.18 & 100.00\\
Random Forest & 16/16 & 69.14 & 5.79 & 5.25 & 3.37 & 100.00\\
LightGBM (leaf 4) & 16/16 & 82.94 & 6.39 & 4.90 & 1.50 & 100.00\\
LightGBM (leaf 20) & 16/16 & 659.16 & 65.42 & 46.18 & 27.03 & 50.00\\
NN lookup & 16/16 & 139.59 & 10.22 & 8.51 & 6.02 & 100.00\\
Qwen zero-shot & 16/16 & 1262.69 & 98.37 & 98.94 & -98.37 & 0.00\\
Qwen five-shot & 16/16 & 518.20 & 50.20 & 35.66 & 14.10 & 62.50\\
GPT-OSS zero-shot & 15/16 & 1075.81 & 90.26 & 95.06 & -90.26 & 0.00\\
GPT-OSS five-shot & 16/16 & 812.61 & 109.35 & 36.88 & 89.90 & 81.25\\
Claude zero-shot & 16/16 & 326.07 & 21.65 & 19.68 & 3.27 & 100.00\\
Claude five-shot & 16/16 & 93.61 & 8.38 & 6.76 & -2.46 & 100.00\\
\bottomrule
\end{tabular}
\caption{Controlled collection, development partition (16 configurations). One GPT-OSS zero-shot response is invalid. Coverage and within-factor-two accuracy use all 16 queries; its error
metrics use the 15 valid predictions. Errors are computed over valid predictions, and within-2$\times$ accuracy counts failures as misses. Median APE and signed RE are percentages.}
\label{tab:development}
\end{table*}
\begin{table*}[tp]
\centering\small
\begin{tabular}{lrrrrrr}
\toprule
Method & Coverage & MAE (ms) & MAPE (\%) & Med. APE & Signed RE & \shortstack{Within\\factor 2 (\%)}\\
\midrule
Median & 8/8 & 879.91 & 38.96 & 42.18 & -38.96 & 50.00\\
OLS & 8/8 & 127.04 & 6.49 & 4.49 & 0.88 & 100.00\\
Ridge & 8/8 & 129.57 & 6.48 & 4.45 & -0.82 & 100.00\\
Random Forest & 8/8 & 92.83 & 5.04 & 3.55 & 1.17 & 100.00\\
LightGBM (leaf 4) & 8/8 & 106.02 & 5.15 & 4.13 & 2.42 & 100.00\\
LightGBM (leaf 20) & 8/8 & 891.42 & 39.62 & 42.80 & -39.62 & 50.00\\
NN lookup & 8/8 & 110.77 & 5.85 & 5.56 & -2.85 & 100.00\\
Qwen zero-shot & 8/8 & 1942.43 & 99.34 & 99.33 & -99.34 & 0.00\\
Qwen five-shot & 8/8 & 716.57 & 32.30 & 29.09 & -17.76 & 75.00\\
GPT-OSS zero-shot & 8/8 & 1783.54 & 90.79 & 89.83 & -90.79 & 0.00\\
GPT-OSS five-shot & 8/8 & 740.10 & 33.32 & 21.15 & 13.35 & 75.00\\
Claude zero-shot & 8/8 & 452.42 & 23.39 & 21.34 & 23.39 & 100.00\\
Claude five-shot & 8/8 & 107.84 & 5.10 & 4.84 & -2.68 & 100.00\\
\bottomrule
\end{tabular}
\caption{Controlled holdout, Qwen2.5-0.5B configurations only (eight configurations; exploratory). Errors are computed over valid predictions, and within-2$\times$ accuracy counts failures as misses. Median APE and signed RE are percentages.}
\label{tab:slice-qwen2.5-0.5b}
\end{table*}
\begin{table*}[tp]
\centering\small
\begin{tabular}{lrrrrrr}
\toprule
Method & Coverage & MAE (ms) & MAPE (\%) & Med. APE & Signed RE & \shortstack{Within\\factor 2 (\%)}\\
\midrule
Median & 8/8 & 454.57 & 89.61 & 82.89 & 89.61 & 50.00\\
OLS & 8/8 & 34.64 & 5.44 & 4.65 & 2.31 & 100.00\\
Ridge & 8/8 & 36.95 & 5.92 & 4.73 & 4.06 & 100.00\\
Random Forest & 8/8 & 26.99 & 4.90 & 4.09 & 1.79 & 100.00\\
LightGBM (leaf 4) & 8/8 & 23.24 & 4.06 & 4.17 & 3.76 & 100.00\\
LightGBM (leaf 20) & 8/8 & 443.07 & 87.58 & 80.93 & 87.58 & 50.00\\
NN lookup & 8/8 & 87.32 & 12.65 & 11.24 & -0.87 & 100.00\\
Qwen zero-shot & 8/8 & 607.91 & 97.95 & 97.95 & -97.95 & 0.00\\
Qwen five-shot & 8/8 & 415.60 & 79.35 & 51.00 & 58.72 & 62.50\\
GPT-OSS zero-shot & 8/8 & 549.93 & 89.00 & 93.31 & -89.00 & 0.00\\
GPT-OSS five-shot & 8/8 & 446.78 & 81.09 & 60.11 & 53.00 & 62.50\\
Claude zero-shot & 8/8 & 90.37 & 14.56 & 14.78 & -5.79 & 100.00\\
Claude five-shot & 8/8 & 69.81 & 11.35 & 10.28 & -6.51 & 100.00\\
\bottomrule
\end{tabular}
\caption{Controlled holdout, T5-small configurations only (eight configurations; exploratory). Errors are computed over valid predictions, and within-2$\times$ accuracy counts failures as misses. Median APE and signed RE are percentages.}
\label{tab:slice-t5-small}
\end{table*}

%% file: generated_analytical.tex
\begin{table*}[tp]
\centering\small
\begin{tabular}{lrrrrrr}
\toprule
Method & Coverage & MAE (ms) & MAPE (\%) & Med. APE & Signed RE & \shortstack{Within\\factor 2 (\%)}\\
\midrule
Direct LightGBM & 42/42 & 11.91 & 18.08 & 13.48 & 3.83 & 100.00\\
Analytical estimate & 42/42 & 22.81 & 56.04 & 60.29 & -56.04 & 26.19\\
Learned correction & 42/42 & 2.75 & 14.45 & 6.78 & 3.62 & 100.00\\
Stacking & 42/42 & 5.47 & 12.07 & 8.55 & 3.31 & 100.00\\
\bottomrule
\end{tabular}
\caption{FP32 forward-pass study: 42 dev runs (40 text encoders and two ViT runs). All learned methods receive the same static features. Median APE and signed RE are percentages.}
\label{tab:fp32}
\par\medskip
\begin{tabular}{lrrrrrr}
\toprule
Method & Coverage & MAE (ms) & MAPE (\%) & Med. APE & Signed RE & \shortstack{Within\\factor 2 (\%)}\\
\midrule
Direct LightGBM & 40/40 & 12.10 & 18.14 & 13.48 & 3.50 & 100.00\\
Analytical estimate & 40/40 & 22.43 & 56.56 & 60.29 & -56.56 & 25.00\\
Learned correction & 40/40 & 2.81 & 15.02 & 6.99 & 3.87 & 100.00\\
Stacking & 40/40 & 5.67 & 12.52 & 8.99 & 3.41 & 100.00\\
\bottomrule
\end{tabular}
\caption{FP32 forward-pass study, text-encoder runs only (40).}
\label{tab:fp32-transformer_encoder}
\par\medskip
\begin{tabular}{lrrrrrr}
\toprule
Method & Coverage & MAE (ms) & MAPE (\%) & Med. APE & Signed RE & \shortstack{Within\\factor 2 (\%)}\\
\midrule
Direct LightGBM & 2/2 & 8.12 & 16.92 & 16.92 & 10.45 & 100.00\\
Analytical estimate & 2/2 & 30.42 & 45.52 & 45.52 & -45.52 & 50.00\\
Learned correction & 2/2 & 1.57 & 3.01 & 3.01 & -1.19 & 100.00\\
Stacking & 2/2 & 1.55 & 3.06 & 3.06 & 1.46 & 100.00\\
\bottomrule
\end{tabular}
\caption{FP32 forward-pass study, ViT runs only (two; descriptive).}
\label{tab:fp32-vision_transformer}
\end{table*}

%% file: analytical_contract.tex
\subsection{Analytical model specification}

Let $B,S,H,I,A,L$ denote batch size, token count, hidden
size, feed-forward size, number of attention heads, and
number of layers. Counting a multiply--add as two FLOPs
and using $q=4$ bytes per value, the transformer blocks
contribute $F_b$ FLOPs and $D_b$ bytes:
\begin{align}
 F_b &= L(8BSH^2+4BSHI+4BS^2H),\\
 D_b &= qL[BS(14H+2I)+4H^2\nonumber\\
 &\qquad +2HI+2BAS^2].
\end{align}

Encoder embeddings add $2qBSH+16BS$ bytes.
BERT and RoBERTa poolers add $2BH^2$ FLOPs and
$q(2BH+H^2)$ bytes; DistilBERT has no pooler.
All text encoders use $H=768$, $I=3072$, and $A=12$,
with $L=12$ for BERT and RoBERTa and $L=6$ for DistilBERT.

ViT uses the same $H$, $I$, and $A$, with 12 layers,
196 patches plus a class token from a $224\times224$
three-channel image, patch size 16, and a 1,000-class head.
Patch embedding and the classification head add
\begin{align}
 F_{\mathrm{extra}}
 &= 2B\cdot196\cdot768\cdot3\cdot16^2 \nonumber\\
 &\quad +2B\cdot768\cdot1000
\end{align}
FLOPs and
\begin{align}
 D_{\mathrm{extra}}
 &= 4\bigl[B(3\cdot224^2+196\cdot768) \nonumber\\
 &\quad +768\cdot3\cdot16^2 \nonumber\\
 &\quad +B(768+1000)+768\cdot1000\bigr]
\end{align}
bytes.

The model assumes one weight read per batched operator
and materialized attention without reuse across operators.
It omits softmax, normalization, nonlinearities, and other
unlisted work, and uses no profiler measurements.

\paragraph{Hardware constants and runtime estimate.}
The nominal GPU peaks are A100 $(19.5,1555)$,
A40 $(37.4,696)$, and T4 $(8.1,320)$, in FP32 TFLOP/s
and GB/s. The retained A100 configuration is labeled SXM,
whereas the run metadata identifies PCIe hardware.
This discrepancy affects 36 fitting and 15 development
records. We retain the original constants, so these results
evaluate the specified nominal scenario rather than a
hardware-verified A100 model.

With $F$ the total modeled FLOPs and $D$ the total modeled
bytes, including the applicable embedding and head terms,
the estimate in milliseconds is
\begin{equation}
 T_{\mathrm{ana,ms}}=1000\max(F/P,D/W),
\end{equation}
where $P$ and $W$ are the nominal FLOP/s and bytes/s:
the values above multiplied by $10^{12}$ and $10^9$,
respectively. The maximum is taken after summing over
operators, with efficiency 1 and no added overhead.
Neither is calibrated on development data.

\paragraph{Predictor inputs.}
The direct and correction models receive the same six
forward-pass inputs: batch size, GPU type, model, precision,
text length (zero for vision), and a text-length
applicability flag.

Both also receive 19 static features: number of layers,
hidden and feed-forward sizes, number of heads, pooler
presence, class count, image height, width, and channels,
patch size, a vision indicator, token count, nominal compute
and bandwidth, bytes per value, modeled FLOPs and bytes,
log analytical milliseconds, and the hardware scenario.
Non-applicable static fields are encoded as zero;
the vision indicator distinguishes image workloads.
Both models use preprocessing fit only on fitting records
and the same fixed tree capacity.

\paragraph{Correction and stacking.}
For a fitting label $y$ in milliseconds, the direct LightGBM
model learns $\log y$, and the correction model learns
$r=\log y-\log T_{\mathrm{ana,ms}}$.
With $c(z)=\min(30,\max(-30,z))$, the predictions are
\begin{align}
 \hat y_{\mathrm{dir}}
 &= \exp(c(f_{\mathrm{dir}}(x))),\\
 \hat y_{\mathrm{corr}}
 &= \exp\!\left(c\!\left(
 \log T_{\mathrm{ana,ms}}+f_{\mathrm{corr}}(x)
 \right)\right).
\end{align}
Both use the default LightGBM settings of
Appendix~\ref{app:settings}: 200 rounds and minimum leaf
size 20.

The stacker is a Ridge regression with penalty 1 and an
unpenalized intercept. Its inputs are
$[1,\log\hat y_{\mathrm{dir}},\log\hat y_{\mathrm{corr}}]$,
and its target is $\log y$.
Fitting groups are sorted by the SHA-256 hash of
\texttt{["baseline-oof-v1", group]} and assigned cyclically
to three folds. For each fold, preprocessing and both base
models are fit on the other two folds. Their clipped
predictions on the held-out fold form the stacker's
training rows.

The base models are then refit on all fitting groups.
At prediction time, the stacker combines their log
predictions and returns $\exp(c(z_{\mathrm{stack}}))$.
Development labels are used neither in the internal folds
nor in fitting the stacker.

%% file: generated_pairs.tex
\begin{table*}[tp]
\begin{minipage}{\textwidth}
\section{Paired Comparisons on the Controlled Holdout}
\label{app:pairs}

We report all 78 pairwise comparisons among the 13 methods,
using the same 16 configurations for each comparison.
Differences are computed as the first method minus the
second; negative values favor the first method.
An interval containing zero does not establish a clear
ordering or demonstrate equivalence.
\end{minipage}
\par
\centering\small
\begin{tabular}{lrrr}
\toprule
Contrast & Common $n$ & $\Delta$MAE (ms), 95\% CI
& $\Delta$MAPE (pp), 95\% CI\\
\midrule
OLS $-$ Median & 16 & [-827.28, -366.57] & [-84.69, -35.46]\\
Ridge $-$ Median & 16 & [-822.54, -366.96] & [-84.52, -35.46]\\
Ridge $-$ OLS & 16 & [-13.15, 17.22] & [-0.64, 1.08]\\
Random Forest $-$ Median & 16 & [-849.59, -384.12] & [-84.44, -36.99]\\
Random Forest $-$ OLS & 16 & [-83.89, 35.33] & [-4.53, 2.32]\\
Random Forest $-$ Ridge & 16 & [-88.31, 33.84] & [-4.90, 2.24]\\
LightGBM (leaf 20) $-$ Median & 16 & [-5.75, 5.75] & [-1.40, 0.00]\\
LightGBM (leaf 20) $-$ OLS & 16 & [365.08, 830.17] & [35.28, 83.43]\\
LightGBM (leaf 20) $-$ Ridge & 16 & [364.50, 824.60] & [35.14, 83.23]\\
LightGBM (leaf 20) $-$ Random Forest & 16 & [382.74, 851.86] & [36.78, 83.28]\\
LightGBM (leaf 4) $-$ Median & 16 & [-836.20, -390.27] & [-84.98, -37.25]\\
LightGBM (leaf 4) $-$ OLS & 16 & [-80.00, 45.60] & [-4.62, 1.74]\\
LightGBM (leaf 4) $-$ Ridge & 16 & [-84.58, 40.48] & [-4.95, 1.58]\\
LightGBM (leaf 4) $-$ Random Forest & 16 & [-18.08, 30.04] & [-2.07, 1.35]\\
LightGBM (leaf 4) $-$ LightGBM (leaf 20) & 16 & [-839.40, -388.39] & [-83.72, -37.08]\\
NN lookup $-$ Median & 16 & [-819.98, -335.56] & [-81.75, -31.78]\\
NN lookup $-$ OLS & 16 & [-32.43, 61.54] & [-0.25, 6.66]\\
NN lookup $-$ Ridge & 16 & [-36.26, 58.27] & [-0.55, 6.51]\\
NN lookup $-$ Random Forest & 16 & [-3.47, 77.88] & [0.62, 8.40]\\
NN lookup $-$ LightGBM (leaf 20) & 16 & [-821.85, -334.12] & [-80.50, -31.46]\\
NN lookup $-$ LightGBM (leaf 4) & 16 & [-14.58, 77.17] & [0.88, 8.99]\\
Qwen zero-shot $-$ Median & 16 & [346.09, 853.67] & [8.25, 57.16]\\
Qwen zero-shot $-$ OLS & 16 & [840.34, 1574.20] & [90.24, 94.81]\\
Qwen zero-shot $-$ Ridge & 16 & [840.55, 1572.51] & [89.90, 94.76]\\
Qwen zero-shot $-$ Random Forest & 16 & [852.10, 1603.40] & [91.94, 95.32]\\
Qwen zero-shot $-$ LightGBM (leaf 20) & 16 & [350.55, 849.32] & [9.46, 57.40]\\
\bottomrule
\end{tabular}
\caption{Paired error differences, part 1 of 3.
CI: confidence interval; pp: percentage points.
Intervals use 10,000 paired bootstrap samples of the
16 configurations and are not adjusted for multiple comparisons.
Endpoints are rounded to two decimal places.}
\label{tab:pairs-1}
\end{table*}

\begin{table*}[tp]
\centering\small
\begin{tabular}{lrrr}
\toprule
Contrast & Common $n$ & $\Delta$MAE (ms), 95\% CI
& $\Delta$MAPE (pp), 95\% CI\\
\midrule
Qwen zero-shot $-$ LightGBM (leaf 4) & 16 & [854.50, 1587.58] & [92.19, 95.82]\\
Qwen zero-shot $-$ NN lookup & 16 & [810.53, 1571.31] & [85.86, 92.33]\\
Qwen five-shot $-$ Median & 16 & [-261.85, 63.28] & [-37.11, 20.74]\\
Qwen five-shot $-$ OLS & 16 & [270.97, 725.24] & [23.86, 84.88]\\
Qwen five-shot $-$ Ridge & 16 & [272.34, 715.21] & [24.16, 84.16]\\
Qwen five-shot $-$ Random Forest & 16 & [276.85, 761.43] & [24.52, 86.18]\\
Qwen five-shot $-$ LightGBM (leaf 20) & 16 & [-262.52, 63.88] & [-36.03, 21.54]\\
Qwen five-shot $-$ LightGBM (leaf 4) & 16 & [277.73, 746.23] & [24.76, 86.79]\\
Qwen five-shot $-$ NN lookup & 16 & [240.42, 716.31] & [20.05, 81.83]\\
Qwen five-shot $-$ Qwen zero-shot & 16 & [-1021.38, -374.27] & [-69.03, -7.27]\\
GPT-OSS zero-shot $-$ Median & 16 & [256.88, 726.63] & [-0.28, 48.50]\\
GPT-OSS zero-shot $-$ OLS & 16 & [755.09, 1446.97] & [78.77, 88.56]\\
GPT-OSS zero-shot $-$ Ridge & 16 & [753.45, 1441.47] & [78.32, 88.45]\\
GPT-OSS zero-shot $-$ Random Forest & 16 & [769.19, 1475.34] & [80.56, 88.67]\\
GPT-OSS zero-shot $-$ LightGBM (leaf 20) & 16 & [260.87, 722.01] & [1.01, 48.69]\\
GPT-OSS zero-shot $-$ LightGBM (leaf 4) & 16 & [769.47, 1462.41] & [80.35, 89.66]\\
GPT-OSS zero-shot $-$ NN lookup & 16 & [723.90, 1440.47] & [74.92, 85.70]\\
GPT-OSS zero-shot $-$ Qwen zero-shot & 16 & [-161.84, -62.20] & [-13.28, -4.87]\\
GPT-OSS zero-shot $-$ Qwen five-shot & 16 & [279.84, 889.49] & [-4.43, 61.53]\\
GPT-OSS five-shot $-$ Median & 16 & [-364.71, 191.53] & [-31.73, 13.08]\\
GPT-OSS five-shot $-$ OLS & 16 & [224.17, 887.90] & [27.58, 78.37]\\
GPT-OSS five-shot $-$ Ridge & 16 & [227.76, 877.17] & [27.79, 77.62]\\
GPT-OSS five-shot $-$ Random Forest & 16 & [268.22, 895.55] & [29.30, 78.22]\\
GPT-OSS five-shot $-$ LightGBM (leaf 20) & 16 & [-366.58, 191.18] & [-30.90, 13.89]\\
GPT-OSS five-shot $-$ LightGBM (leaf 4) & 16 & [276.16, 871.70] & [29.97, 78.87]\\
GPT-OSS five-shot $-$ NN lookup & 16 & [215.12, 858.48] & [24.58, 74.99]\\
\bottomrule
\end{tabular}
\caption{Paired error differences, part 2 of 3.
Definitions and interval construction follow
Table~\ref{tab:pairs-1}.}
\label{tab:pairs-2}
\end{table*}

\begin{table*}[tp]
\centering\small
\begin{tabular}{lrrr}
\toprule
Contrast & Common $n$ & $\Delta$MAE (ms), 95\% CI
& $\Delta$MAPE (pp), 95\% CI\\
\midrule
GPT-OSS five-shot $-$ Qwen zero-shot & 16 & [-1084.46, -322.49] & [-64.78, -14.60]\\
GPT-OSS five-shot $-$ Qwen five-shot & 16 & [-285.52, 331.01] & [-31.43, 26.38]\\
GPT-OSS five-shot $-$ GPT-OSS zero-shot & 16 & [-944.29, -244.06] & [-55.97, -6.36]\\
Claude zero-shot $-$ Median & 16 & [-599.32, -192.31] & [-72.54, -20.55]\\
Claude zero-shot $-$ OLS & 16 & [68.07, 327.82] & [6.87, 18.88]\\
Claude zero-shot $-$ Ridge & 16 & [60.38, 332.90] & [6.18, 19.13]\\
Claude zero-shot $-$ Random Forest & 16 & [97.43, 352.02] & [8.38, 20.87]\\
Claude zero-shot $-$ LightGBM (leaf 20) & 16 & [-600.31, -191.65] & [-71.28, -20.48]\\
Claude zero-shot $-$ LightGBM (leaf 4) & 16 & [103.52, 338.33] & [9.16, 20.35]\\
Claude zero-shot $-$ NN lookup & 16 & [48.97, 317.81] & [2.10, 17.65]\\
Claude zero-shot $-$ Qwen zero-shot & 16 & [-1322.95, -713.56] & [-85.12, -73.29]\\
Claude zero-shot $-$ Qwen five-shot & 16 & [-541.59, -71.29] & [-73.98, -8.34]\\
Claude zero-shot $-$ GPT-OSS zero-shot & 16 & [-1201.04, -621.81] & [-77.84, -63.19]\\
Claude zero-shot $-$ GPT-OSS five-shot & 16 & [-661.61, -34.86] & [-65.78, -14.12]\\
Claude five-shot $-$ Median & 16 & [-818.47, -360.14] & [-81.00, -34.40]\\
Claude five-shot $-$ OLS & 16 & [-33.60, 48.14] & [-1.63, 6.42]\\
Claude five-shot $-$ Ridge & 16 & [-34.90, 40.92] & [-1.89, 6.11]\\
Claude five-shot $-$ Random Forest & 16 & [-21.33, 73.40] & [-0.26, 6.74]\\
Claude five-shot $-$ LightGBM (leaf 20) & 16 & [-820.97, -357.65] & [-79.80, -34.04]\\
Claude five-shot $-$ LightGBM (leaf 4) & 16 & [-28.33, 69.73] & [0.06, 7.32]\\
Claude five-shot $-$ NN lookup & 16 & [-42.34, 20.11] & [-4.97, 2.80]\\
Claude five-shot $-$ Qwen zero-shot & 16 & [-1573.10, -827.65] & [-93.43, -86.90]\\
Claude five-shot $-$ Qwen five-shot & 16 & [-715.51, -263.35] & [-82.61, -21.99]\\
Claude five-shot $-$ GPT-OSS zero-shot & 16 & [-1440.83, -741.27] & [-86.00, -76.96]\\
Claude five-shot $-$ GPT-OSS five-shot & 16 & [-857.75, -229.97] & [-73.82, -27.14]\\
Claude five-shot $-$ Claude zero-shot & 16 & [-326.04, -60.86] & [-18.53, -3.60]\\
\bottomrule
\end{tabular}
\caption{Paired error differences, part 3 of 3.
Definitions and interval construction follow
Table~\ref{tab:pairs-1}.}
\label{tab:pairs-3}
\end{table*}

%% file: technical_specifications.tex
\section{Input Fields, Lookup, and Splits}
\label{app:schemas}

The companion file \texttt{TECHNICAL\_SPECIFICATIONS.json}
lists exact field names and source hashes.
Identifiers, split assignments, paths, hashes, measured
resource use, and execution outcomes are excluded from
predictor inputs. Runtime labels are stored separately,
and each timing target has its own preprocessing and model.

\paragraph{Historical training (30 fields).}
Model and task fields are model, dataset, task type,
input modality, framework, and preprocessing.
Execution fields are GPU type, requested GPU count,
parallelism, precision, quantization, adapter mode,
gradient checkpointing, and batch semantics.
Workload and optimization fields are batch size, epochs,
sequence length, image resolution, frame count, maximum
steps, execution extent, step semantics, gradient
accumulation, LoRA rank and alpha, optimizer, learning
rate, learning-rate schedule, and number of data-loader
workers. The timing procedure is an explicit categorical
field.

Values come from literal launch arguments, fixed dispatch
in the script, or script defaults, never from measured
outcomes. Non-applicable numeric fields use zero; existing
fields such as input modality and adapter mode distinguish
their applicability. There is no separate applicability
indicator for every field. A maximum-step value of zero
means no cap. Unresolved Trainer worker and checkpointing
defaults are represented by named unknown categories.

Numeric columns are batch size, epochs, gradient
accumulation, image resolution, learning rate, LoRA rank
and alpha, maximum steps, frame count, GPU count, and
sequence length. Worker count is numeric for custom loops
and categorical for Trainer runs. All other fields,
including booleans, are categorical.

\paragraph{Historical inference.}
The forward-pass target uses six fields: batch size,
GPU type, model, precision, text length, and a flag
indicating whether text length applies. Text length is
zero for vision models.

The other inference targets use seven fields: batch size,
GPU type, input length, model, precision, requested output
limit, and the product of batch size and input length.
The analytical study adds the 19 static features listed
in Appendix~\ref{app:analytical}.

\paragraph{Controlled generation (15 fields).}
Workload fields are model, task, batch size, and input
length. Decoding fields are requested output limit,
number of beams, number of returned sequences, sampling
flag, cache flag, and seed. Execution fields are precision,
attention implementation, TF32 flag, GPU model, and
requested GPU count.

All fields are required, and constant fields remain
explicit. Realized output length, measured memory use,
and latency are excluded from predictor inputs.

\paragraph{Encoding.}
Columns are ordered lexically. For schemas without fixed
column types, a column is numeric if all known fitting
values are non-boolean numbers and categorical otherwise.
Categorical vocabularies are sorted and include an unknown
bucket. Numeric preprocessing and regression settings
follow Appendix~\ref{app:settings}. Preprocessing is fit
only on fitting records and is never refit on development
or holdout queries.

\paragraph{NN lookup.}
Distances are Euclidean in the standardized and one-hot
encoded fitting matrix. For controlled generation,
candidates must match the query's model, GPU, and precision
exactly. The prediction is the median runtime of candidates
tied at the minimum distance; lookup abstains when no
compatible candidate exists. It returns observed fitting
runtimes without scaling them to the query's workload size.

Historical inference lookup additionally requires the
timing procedure and the reference implementation's task,
strategy, and execution-extent compatibility fields.
Forward-sum records lack some required compatibility
information, so lookup abstains on these targets even
when model, GPU, and precision match.

Historical training lookup requires agreement on model,
GPU, precision, task, adapter, dataset, execution extent,
batch semantics, gradient accumulation, and parallelism.
It first takes the median runtime of records with identical
feature vectors, then returns the median over the nearest
such entries. Ties use relative and absolute distance
tolerances of $10^{-12}$.

NN lookup + LightGBM uses the fixed LightGBM predictor
whenever lookup abstains. The controlled comparison
includes standalone NN lookup only.

\paragraph{Historical grouping and splits.}
Inference grouping uses normalized configuration fields:
model and revision; GPU type and count; precision;
task and architecture; dataset and revision; image or
input shape; frames; input and output lengths; batch size
and semantics; parallelism and tensor or pipeline degree;
quantization and cache data type; decoding, beam, sampling,
and stopping settings; serving framework, batching, and
concurrency; adapters; preprocessing; and available code
or environment fingerprints.

Within a timing target, records are linked when every
field known for both agrees. Connected components of
these links define configuration groups.
Groups are ranked by the SHA-256 hash of
\texttt{[seed, procedure, group]}, with seed
\texttt{physboost-primary-v1}. Approximately 60\%, 20\%,
and 20\% of groups are assigned to fitting, development,
and the original test partition, with at least one group
per partition when possible. These percentages apply
to groups, not individual records. The original inference
test partition was examined during earlier development
and is not evaluated in this paper.

Training records are grouped by a hash of all 30 input
fields, including the timing procedure. Within each
procedure, groups are ranked by the SHA-256 hash of
\texttt{[training-development-v1, 0, procedure, group]}.
The first 20\% of groups, rounded up, go to development
and the rest to fitting, subject to a minimum fitting
size. We retain the existing grouping and assignments.

\paragraph{Controlled splits.}
For each workload model, batch sizes $1,2,4,8$ are indexed
by $b=0,\ldots,3$, input lengths $128,512$ by $i=0,1$,
output limits $64,128$ by $o=0,1$, and greedy or four-beam
decoding by $d=0,1$. We compute
\begin{equation}
v=(4i+2o+d+2b)\bmod 8.
\end{equation}
Configurations with $v\in\{0,3,5,6\}$ go to fitting,
$v\in\{1,4\}$ to development, and $v\in\{2,7\}$ to the
holdout. This gives 16 fitting, eight development,
and eight holdout configurations per workload model.

The assignment rule is deterministic and uses no random
seed; the token-bank seed is a separate input setting.
Configurations sharing the pilot settings are outside
the controlled holdout.

\paragraph{LLM output formats.}
Historical requests specify their timing target and unit
and require a positive \texttt{predicted\_runtime}.
Controlled requests require \texttt{prediction\_ms}
with unit \texttt{ms\_per\_batch}.
Both use the fixed parsing rules described in
Appendix~\ref{app:settings}.